\ifdefined\pdfminorversion
\fi
\documentclass[]{beingbeyond}
\usepackage[toc,page,header]{appendix}

\usepackage[utf8]{inputenc} %
\usepackage[T1]{fontenc}    %
\usepackage{hyperref}       %
\usepackage{url}            %
\usepackage{array}          %
\usepackage{booktabs}       %
\usepackage{amsfonts}       %
\usepackage{nicefrac}       %
\usepackage{microtype}      %
\usepackage{xcolor}         %
\usepackage{xspace}
\usepackage{bm}
\usepackage{tabularx}
\usepackage{amssymb}
\usepackage{enumitem}
\usepackage{amsmath}
\usepackage{mathtools}
\usepackage{amsthm}
\usepackage{multirow}
\usepackage{makecell}
\usepackage{color}
\usepackage{colortbl}
\usepackage{adjustbox}
\usepackage{graphicx}
\usepackage{soul}
\usepackage{pifont}
\usepackage{wrapfig}
\usepackage{multicol}
\usepackage{tikz}

\definecolor{methodPsiZero}{HTML}{FCE8E6}
\definecolor{methodGroot}{HTML}{FFC6BC}
\definecolor{methodPiZeroFive}{HTML}{F8B9B8}
\definecolor{methodInternVLA}{HTML}{D6DFEF}
\definecolor{methodHRDT}{HTML}{A5CDE2}
\definecolor{methodDreamZero}{HTML}{5FA3CB}
\definecolor{methodEgoVLA}{HTML}{668FCA}
\definecolor{methodDP}{HTML}{C9CEFE}
\definecolor{methodACT}{HTML}{66A8CD}
\definecolor{methodBeing}{HTML}{006BAD}

\makeatletter
\DeclareRobustCommand\onedot{\futurelet\@let@token\@onedot}
\def\@onedot{\ifx\@let@token.\else.\null\fi\xspace}

\makeatother
\graphicspath{{fig/}}

\definecolor{BlockC}{gray}{0.98}
\definecolor{BlockA}{RGB}{191,211,230}
\definecolor{BlockB}{RGB}{199,233,192}

\definecolor{cgreen}{RGB}{0, 150, 0}
\definecolor{cred}{RGB}{200, 0, 0}
\definecolor{cblue}{RGB}{0, 0, 150}

\title{Being-M0.7: A Latent World-Action Model \\ for Humanoid Robots}

\author{\textbf{BeingBeyond Team}}
\webpage{\url{https://research.beingbeyond.com/being-m07}}

\hypersetup{
  pdftitle={Being-M0.7: A Latent World-Action Model for Humanoid Robots},
  pdfauthor={BeingBeyond Team}
}

\firstfig[width=\linewidth][1.0\textwidth]
{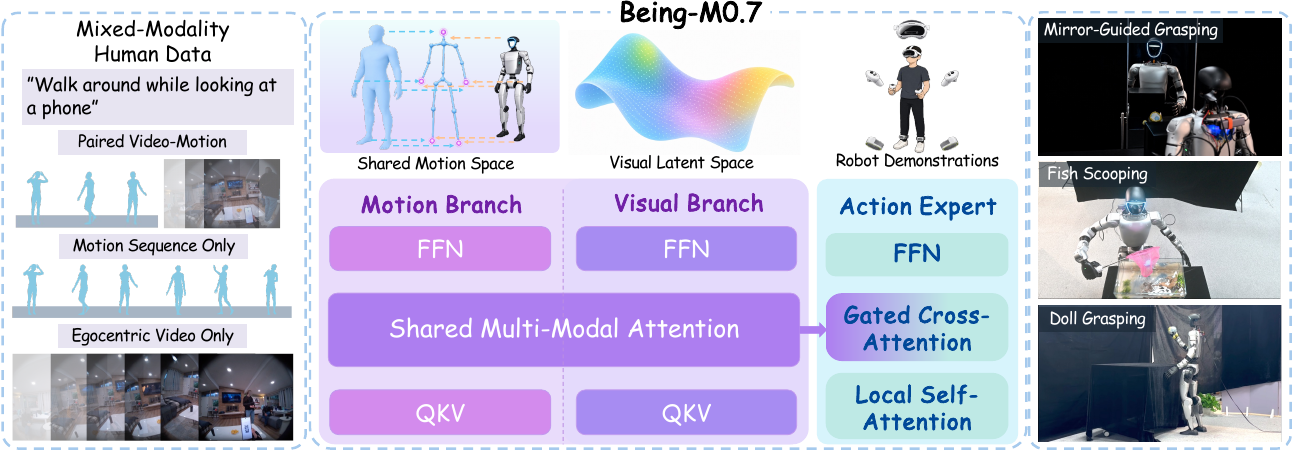}
{\textbf{Being-M0.7 at a glance.} Being-M0.7 is a latent World-Action Model pretrained on egocentric videos and human motions with a Mixture of Transformers (MoT) structure, then grounded into humanoid control through action expert post-training on robot trajectories for diverse loco-manipulation tasks. The rightmost panels show three real-world tasks executed by the humanoid.}
{fig:first_fig_label}

\abstract{
Humanoid loco-manipulation requires coordinated locomotion and manipulation informed by future scene evolution and whole-body motion, yet learning these capabilities is constrained by scarce robot demonstrations.
Human video and motion datasets offer scalable supervision, but many contain only video or motion rather than paired video-motion data. Moreover, human motion does not directly specify executable robot actions.
We present \textbf{Being-M0.7}, a latent world-action model that transfers visual--motion priors learned from mixed-modality human data to humanoid control through pre-training, robot mid-training, and action post-training.
We curate a corpus from more than 10,000 hours of raw human-centric data, integrating video-only, motion-only, and paired video--motion streams to learn complementary visual dynamics and whole-body kinematic structure.
Joint prediction of future latent visual states and motion encourages visual representations to encode future kinematics.
Robot mid-training adapts this coarse-grained prior to robot viewpoints and body dynamics.
During action post-training, an action expert combines visual predictive representations from the frozen, adapted prior with current images and proprioception through gated cross-attention, grounding predictive context in executable whole-body commands.
Being-M0.7 achieves the highest aggregate success rate among the compared baselines on SIMPLE and matches the strongest baseline on real-world Unitree G1 loco-manipulation tasks.

}
\checkdata[Date]{October 8, 2026}

\begin{document}
\maketitle

\section{Introduction}

Humanoid loco-manipulation aims to enable a humanoid robot to move through human-centric environments while performing object-centric tasks with coordinated whole-body motion~\citep{he2025omnih2o,ben2025homie,fu2025humanplus,wei2026psi0, jiang2026wholebodyvla, shi2026egohumanoid}. Unlike fixed-base manipulation, the robot must jointly decide where to move, how to orient its body, how to coordinate its hands and feet, and how to maintain a stable posture while interacting with the scene. This makes the problem intrinsically temporal and whole-body: a useful policy should reason about current observations as well as future scene evolution, body motion, and task progress.

Recent robot foundation models and world-action models have made encouraging progress~\citep{brohan2022rt1,zitkovich2023rt2,o2024open,octo2024,pi0,kim2025openvla,du2023learning,agarwal2025cosmos,wam,h05,h07}, but their extension to humanoid loco-manipulation is constrained by the scarcity of robot demonstrations. Collecting synchronized egocentric video, proprioception, and executable whole-body commands requires demanding teleoperation and is limited by safety, hardware availability, and environment diversity~\citep{he2025omnih2o,ben2025homie,fu2025humanplus}. Human egocentric videos and motion capture data offer a scalable source of supervision for visual dynamics and whole-body behavior~\citep{grauman2022ego4d,nymeria,guo2022generating,snap}. 
However, these sources differ in modality coverage, with many providing only video or motion. Moreover, human motion does not directly specify executable robot actions.
Combining video-only, motion-only, and paired video--motion data can expand the scale of pre-training beyond scarce robot demonstrations. The challenge is to transfer these human visual--motion priors to executable humanoid actions.

To this end, we present \textbf{Being-M0.7}, a latent world-action model for humanoid loco-manipulation, together with a training framework that transfers visual--motion priors learned from mixed-modality human data to executable robot actions. We curate a pre-training corpus from more than 10,000 hours of raw human-centric data, organizing it into video-only, motion-only, and paired video--motion streams. Video-only data provides visual dynamics across diverse interaction contexts, motion-only data supplies whole-body kinematic structure, and paired data links visual evolution to body motion. 
The model learns from the available modalities through latent visual prediction and motion prediction. Joint visual--motion prediction encourages the visual representations to focus on underlying kinematic changes rather than the visual details needed for pixel-level reconstruction, providing the downstream policy with motion-informed predictive context for coordinated whole-body action generation.

This coarse-grained visual--motion prior captures future scene evolution and body motion, but does not directly specify executable robot commands for precise interaction. Moreover, a prior model learned from human data must adapt to robot viewpoints and body dynamics. We therefore use robot mid-training to adapt the visual--motion prior model to robot trajectories, followed by action post-training to learn executable whole-body commands. During post-training, a separate action expert combines visual predictive representations from the frozen, adapted prior model with current images and proprioception at each sampled action query. Through gated cross-attention, the expert learns how much predictive information to incorporate from the prior model's visual hidden states. These visual representations also capture information about future kinematics learned through joint visual--motion modeling. 
Together, these three stages form a framework for transferring scalable human visual--motion learning to closed-loop humanoid loco-manipulation tasks.

Our contributions are summarized as follows:
\begin{itemize}[leftmargin=15pt]
\item We propose Being-M0.7, a latent world-action model for humanoid loco-manipulation, with a unified pre-training, mid-training, and post-training framework. Joint visual--motion pre-training from large-scale human video and motion data encourages visual predictive representations to encode future kinematics, while robot mid-training and gated action post-training adapt the coarse-grained visual--motion prior and ground it in executable whole-body control.
\item We curate a mixed-modality pre-training corpus from more than 10,000 hours of raw human-centric data, integrating video-only, motion-only, and paired video--motion streams. These sources complement costly robot demonstrations with visual and whole-body motion supervision.
\item We evaluate Being-M0.7 on the SIMPLE benchmark and real-world Unitree G1 tasks, achieving the highest aggregate simulation success rate among the compared baselines and matching the strongest baseline on real-world tasks. Ablations support the contributions of mixed-modality pre-training, robot mid-training, and predictive context to downstream control.
\end{itemize}

\section{Related Work}

\noindent\textbf{Humanoid Loco-manipulation.}
Learning-based humanoid control combines whole-body motion priors with task-conditioned interaction. Such interaction requires coordinating locomotion, balance, and manipulation as object geometry and contact conditions change. A useful motion prior must therefore be coupled with perceptual feedback to support task execution in the current scene. Reference-based methods use retargeted demonstrations or generated trajectories to guide coordinated behavior~\citep{fu2025humanplus,zhao2025resmimic,weng2025hdmi}, while reference-free approaches learn from task objectives and perception~\citep{lessmimic,xu2025interact}. Reference guidance supplies a concrete target for whole-body coordination, whereas task-driven learning must discover suitable behavior through the interaction objective. This distinction highlights two complementary needs: retaining useful movement structure and allowing the policy to adjust that structure to the object and task at hand. Structured interaction representations and hierarchical control further support geometric generalization~\citep{lessmimic,yin2025visualmimic}. These approaches motivate learning reusable motion knowledge beyond individual skills. Being-M0.7 learns this knowledge through mixed-modality video--motion pre-training, followed by robot adaptation and action grounding.

\noindent\textbf{Egocentric Motion and Joint Prediction.}
Text-conditioned motion models learn semantic whole-body behavior from large motion corpora~\citep{tevet2022human,jiang2023motiongpt}. Egocentric methods additionally connect motion to the agent's viewpoint: EgoEgo and EgoAllo recover body motion from first-person observations~\citep{li2023egoego,yi2025egoallo}, while UniEgoMotion combines reconstruction, forecasting, and generation~\citep{patel2025uniegomotion}. EgoAgent jointly predicts future states and actions~\citep{chen2025egoagent}, and EgoTwin couples egocentric video generation with head-centered human motion~\citep{xiu2025egotwin}. Together, these directions motivate learning how visual observations evolve with body motion, linking perception with coordinated movement. For downstream control, the value of such prediction lies in providing temporal context for action selection as the interaction unfolds. Transferring this knowledge to a humanoid additionally requires an interface between human motion structure and robot commands, with feedback to accommodate differences in embodiment and execution. Our focus is transferring predictive knowledge to humanoid execution: we jointly learn visual-latent and motion dynamics, use a shared human--humanoid motion representation, and train an action expert on the adapted prior model's visual context.

\noindent\textbf{Humanoid Foundation and World-Action Models.}
GR00T~\citep{gr00tn1}, WholeBodyVLA~\citep{jiang2026wholebodyvla}, and $\Psi_0$~\citep{wei2026psi0} explore generalist humanoid policies using heterogeneous supervision, including human videos and robot trajectories. World-action modeling provides a related route through predictive representations that connect anticipation with control. MotionWAM~\citep{zheng2026motionwam} conditions a humanoid motion model on intermediate features from an egocentric video generator. Being-M0.7 also uses intermediate predictive features, with a recipe centered on latent visual--motion learning from paired and unpaired data, robot mid-training, and gated action post-training. In this formulation, video-only and motion-only samples contribute visual dynamics and kinematic supervision, while paired samples connect the two modalities. Motion thus supplies structure during prior learning without requiring explicit motion generation at deployment. The adapted prior model is frozen during action learning; the action expert combines visual context generated by the adapted prior model with current observations through a learned gate. This separation allows human-centric prediction to be learned without robot action labels, while robot supervision determines how the resulting context informs executable commands. The gate makes this contribution conditional on the expert's current observation-dependent features.

\section{Method}

\begin{figure}[t]
    \centering
    \includegraphics[width=\linewidth]{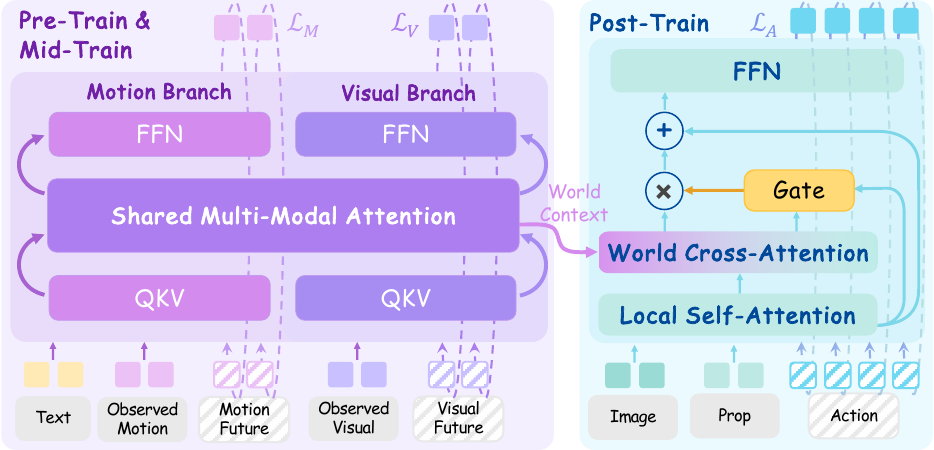}
    \caption{\textbf{Being-M0.7 framework.} Mixed-modality pre-training learns a visual--motion prior with a shared human--humanoid motion representation. Robot mid-training adapts the prior model; gated post-training learns an action expert from its frozen visual context and current robot observations.}
    \label{fig:method}
\end{figure}

Being-M0.7 learns a latent video--motion prior from human-centric data, adapts it to robot trajectories, and then trains a gated action expert against the frozen prior model. We first formulate the problem in~\Cref{sec:problem-formulation}, describe the mixed-modality data and shared representations in~\Cref{sec:data-curation}, and present the architectures of the prior model and action expert in~\Cref{sec:architecture}. 
We then detail the pre-training, robot mid-training, post-training, and closed-loop inference recipes (Figure~\ref{fig:method}).

\subsection{Problem Formulation}
\label{sec:problem-formulation}
Let $V$ be egocentric video, $Z=\mathcal E(V)$ its visual latents, $M$ a compact motion sequence, and $I$ the task instruction. 
Here, $\mathcal E$ is a frozen DINO encoder~\citep{simeoni2025dinov3} whose latents serve as both visual inputs and prediction targets, avoiding pixel-level reconstruction.
Each sample is a temporal window of $T$ consecutive time steps from a video and/or motion sequence, with observed history indexed by $\mathcal C={1,\ldots,K}$ and future targets indexed by $\mathcal Q={K+1,\ldots,T}$. Details appear in Appendix~\ref{app:hyperparameters}.
During pre-training and mid-training, the prior model represents $p_\theta(Z_{\mathcal Q},M_{\mathcal Q}\mid Z_{\mathcal C},M_{\mathcal C},I)$, with motion providing coarse-grained supervision of whole-body dynamics. Joint video--motion modeling encourages the visual future-token representations to capture both visual evolution and the underlying kinematic changes. During post-training and deployment, the prior model generates visual futures from $(Z_{\mathcal C},I)$ with explicit motion prediction disabled. The resulting future-token representations carry learned information about future kinematics and serve as policy conditioning for a future-conditioned action expert~\citep{xie2023future}, which combines this context with the current robot observation $O_q$ to predict fine-grained actions in a chunk $\mathbf a_q=a_{q:q+h-1}$.

\subsection{Data Curation}
\label{sec:data-curation}
\begin{figure}[htbp]
    \centering
    \includegraphics[width=.9\textwidth]{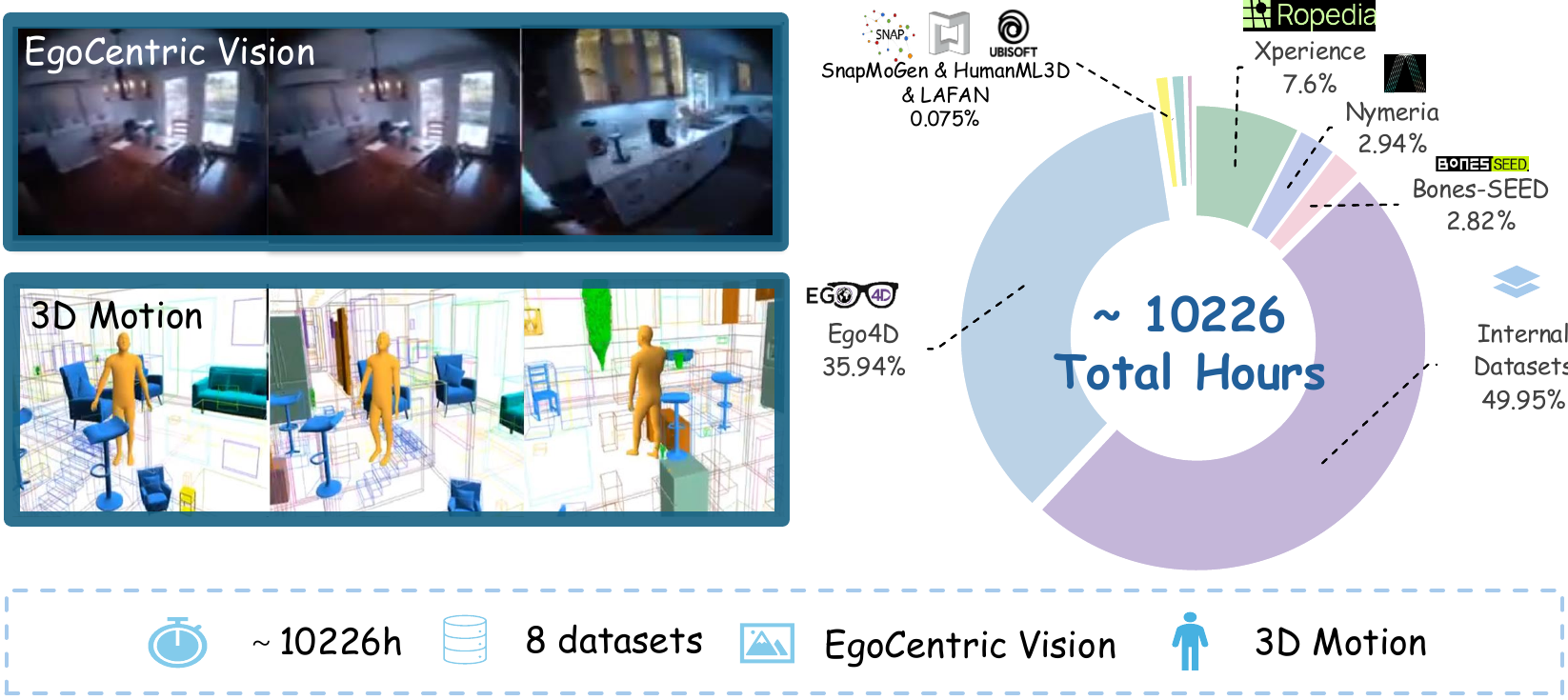}
    \caption{\textbf{Being-M0.7 data recipe.} The pre-training corpus is sourced from external public datasets including Ego4D~\citep{grauman2022ego4d}, Xperience~\citep{xperience}, Nymeria~\citep{nymeria}, Bones-SEED~\citep{luo2025sonic}, SnapMoGen~\citep{snap}, HumanML3D~\citep{guo2022generating}, and Lafan1~\citep{lafan1}, as well as additional data comprising subsets of the training corpora described in Being-H0.5~\citep{h05} and Being-M0.5~\citep{m05, opent2m}, and commercially acquired datasets. Raw sources are filtered, temporally segmented, and reformatted into three supervision streams--$\mathcal{D}_{\textit{VM}}$ for video-motion pairs, $\mathcal{D}_{V}$ for video-only, and $\mathcal{D}_{M}$ for motion-only--which jointly supervise the visual-motion joint distribution and its marginals, respectively.}
    \label{fig:data}
\end{figure}

Our raw collection contains more than 10,000 hours before filtering and temporal segmentation. It yields three supervision streams: paired egocentric video--motion data $\mathcal D_{\textit{VM}}$, video-only data $\mathcal D_V$, and motion-only data $\mathcal D_M$. Paired samples supervise joint prediction; unpaired samples supervise the corresponding modality with shared model parameters. This extends training beyond the paired corpus while retaining cross-modal interaction wherever both modalities are observed. Figure~\ref{fig:data} summarizes the data sources and supervision streams; Appendix~\ref{app:data_details} details motion processing.

\subsection{Model Architecture}
\label{sec:architecture}
Being-M0.7 combines a video--motion prior model with a separate future-conditioned action expert. The expert combines predictive representations learned by the prior model with current robot observations to generate executable commands for downstream robot control tasks.

\paragraph{Video-Motion Mixture of Transformers.}
We implement the prior model as a Mixture of Transformers (MoT)~\citep{liang2024mixtureoftransformers}. Separate input projections, temporal embeddings, and modality embeddings map visual and motion tokens into a common hidden space. Each block retains modality-specific parameters but combines the streams through shared attention, allowing visual-state evolution and motion dynamics to inform one another. 
Both streams cross-attend to task instruction features from a frozen DistilBERT encoder~\citep{sanh2019distilbert}. Modality-specific heads predict flow velocities in the visual latent space and the motion space, respectively.

A chunk attention mask prevents history tokens from accessing future targets, while noisy future tokens attend to the history and to one another. The prior model therefore denoises a complete future chunk in parallel. Only available modalities contribute supervision. Missing-modality tokens are masked out of interactions with valid tokens; Appendix~\ref{app:architecture_details} specifies the token and attention handling. Video-only data supervises visual dynamics across diverse scenes, motion-only data supplies kinematic structure, and paired data connects that structure to visual dynamics. The resulting visual representations provide the downstream expert with motion-informed predictive context without requiring explicit motion generation during control. See Appendices~\ref{app:architecture_details} and~\ref{app:architecture_hyperparameters} for details.

\paragraph{Future-Conditioned Action Expert.}
The action expert is a separate flow-based transformer. It receives noisy action tokens and the current observation $O_q$, including the egocentric image and proprioceptive state. The image is encoded using the same frozen DINO encoder, while proprioception is mapped using a learned projection. Local self-attention first combines action and observation tokens. At selected layers, gated cross-attention then injects hidden states $H^\theta(\tau)$ from the frozen prior model, exposing motion-informed predictive context at multiple network depths. Information flows only from the prior model to the expert: action tokens never enter the prior model's computation.

\paragraph{Gated World Context.}
Let $x_{i,j}$ be action-token $i$ after local observation processing at expert layer $j$, and $r_{i,j}$ its projected world cross-attention update. We gate the injected residual as
\begin{equation}
\begin{aligned}
g_{i,j}&=\sigma\!\left(w_j^\top[\operatorname{LN}(x_{i,j});\operatorname{LN}(r_{i,j})]+b_j\right),\\
\Delta x_{i,j}&=g_{i,j}\bigl(\gamma_j(\tau)\odot r_{i,j}\bigr),
\end{aligned}
\label{eq:world_gate}
\end{equation}
where $\gamma_j(\tau)$ is the flow-conditioned residual scale. The goal of the token-wise gate is to learn how much predictive context to incorporate given the local state and proposed update, allowing the influence of predictive context to vary across tokens. 
The gate controls only the contribution of predictive context; action tokens retain access to current observations through local self-attention.

\subsection{Training and Inference Recipes}
\paragraph{Pre-Training.}
We train visual and motion prediction with flow matching~\citep{lipman2023flowmatching}. For clean future tokens $\bm{x}_0$, sample $\epsilon\sim\mathcal N(0,I)$ and $\tau\sim\mathcal U(0,1)$, and set $\bm{x}_\tau=(1-\tau)\bm{x}_0+\tau\epsilon$. For each modality $k\in\{V,M\}$, we train the model to predict the target flow velocity with the loss
\begin{equation}
\mathcal{L}_k =
\mathbb{E}_{\bm{x}_0^k,\epsilon,\tau}
\left[
\left\|
u_\theta^k(\bm{x}_\tau^k,\tau,c_s)
- (\epsilon-\bm{x}_0^k)
\right\|_2^2
\right],
\label{eq:flow_loss}
\end{equation}
where $c_s$ contains the clean history and task instruction.
Each modality-specific loss is computed only when that modality
is observed. The overall objective is
\begin{equation}
\begin{aligned}
\mathcal L_{\mathrm{pre}}={}&\mathbb E_{\mathcal D_{\textit{VM}}}[\mathcal L_V+\mathcal L_M+\mathcal L_{\mathrm{geom}}]
+\mathbb E_{\mathcal D_V}[\mathcal L_V]\\
&+\mathbb E_{\mathcal D_M}[\mathcal L_M+\mathcal L_{\mathrm{geom}}],
\end{aligned}
\end{equation}
where $\mathcal L_{\mathrm{geom}}$ regularizes global trajectories and local end-effector velocities whenever motion is available; Appendix~\ref{app:geometry_losses} defines both auxiliary terms.

\paragraph{Mid- and Post-Training.}
\label{sec:midtraining}
Both stages use the same robot trajectory dataset, first to adapt the prior model and then to train the action expert. During mid-training, we adapt the prior model to robot video, motion, and task text using $\mathcal L_{\mathrm{mid}}=\mathcal L_V+\mathcal L_M$, where $\mathcal L_M$ is only applied when motion is available. The shared motion representation transfers the pre-training objectives to robot scenes and body dynamics. We then freeze the adapted prior model for post-training with action supervision.

During post-training, we select an action query $q$ in the future interval of each robot trajectory window.
Conditioned on the observed visual history $Z_{\mathcal C}$ and task instruction $I$, the frozen prior model initializes future visual latents from Gaussian noise and integrates its flow ODE to generate a visual future, with explicit motion prediction disabled.
Intermediate hidden states along this denoising trajectory provide predictive context $H^\theta(\tau)$ for the action expert, which also receives the current observation $O_q$.
This trains the expert on the prior model's own predictions, reducing the conditioning mismatch with inference. Further details can be found in Appendix~\ref{app:posttrain}.

Following action chunking~\citep{zhao2023learning}, we corrupt each demonstrated action chunk as
$\tilde{\mathbf a}_{q,\tau}=(1-\tau)\mathbf a_q+\tau\epsilon_a$,
where $\epsilon_a$ is Gaussian noise, and optimize the action predictor with
\begin{equation}
\mathcal L_A=
\mathbb E\left[
\left\|
u_\phi^a(\tilde{\mathbf a}_{q,\tau},\tau,H^\theta(\tau),O_q)
-(\epsilon_a-\mathbf a_q)
\right\|_2^2
\right].
\end{equation}
The frozen latent WAM receives no action-loss gradients in either the visual or motion branch.

\paragraph{Inference.}

The prior model periodically generates a visual-latent plan from recent images and task text. Its intermediate states are cached as policy-side keys and values and reused across action queries. At each query, the expert receives fresh images and proprioception and denoises an action chunk using the gated cache. The cache is refreshed before its context is fully consumed, separating lower-frequency planning from responsive command generation. 
Details are provided in Appendix~\ref{app:inference}.

\section{Experiments}

\subsection{Experimental Setup}

\subsubsection{Hardware Setup}

Our real-world system is built upon a Unitree G1 humanoid robot for whole-body dexterous teleoperation. The platform integrates dual 7-DoF arms with Linker Hand O6 dexterous hands for upper-body manipulation, and an Intel RealSense D435i RGB camera mounted on the robot head to provide egocentric visual observations. Policy inference runs on an external workstation with a single NVIDIA RTX 4090 GPU. 
During deployment, a low-level controller communicates with the policy server via ZMQ in a closed loop and executes predicted commands on the robot in real time.
\begin{figure}[htbp]
    \centering
    \includegraphics[width=0.9\textwidth]{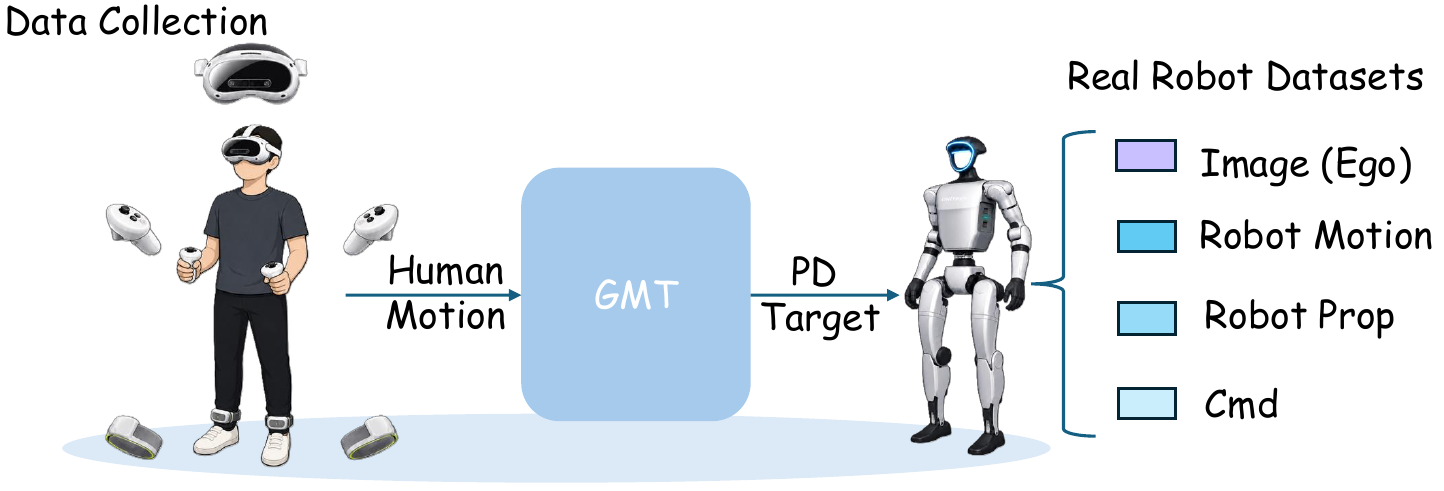}  %
    \caption{\textbf{Overview of the data collection system.} During the data collection stage, an operator uses a VR-based teleoperation system to generate low-level motion commands, and a pretrained whole-body motion tracker controls the humanoid robot. We record synchronized real-world trajectories comprising egocentric images, proprioceptive observations, and motion commands.}
    \label{fig:whole_system_setup}
\end{figure}
\subsubsection{VR-Based Whole-Body Teleoperation Pipeline}

We collect real-robot demonstrations using a VR-based whole-body teleoperation system. The operator wears a PICO VR headset and two ankle-mounted PICO trackers and holds two controllers to provide full-body motion input. XRoboToolkit ~\citep{zhao2026xrobotoolkit} estimates the operator's SMPL pose in real time from the VR device streams, while SONIC ~\citep{luo2025sonic} converts the estimated pose into stable 29-DoF whole-body control commands executed on the Unitree G1 at 50 Hz. During teleoperation, synchronized robot proprioception, egocentric RGB observations, and motion commands (including hand control signals and SONIC motion tokens) are recorded to construct the multi-task post-training dataset. Figure~\ref{fig:whole_system_setup} provides an overview of the proposed system.

\subsubsection{Baselines and Multi-Task Training}

For simulation and real-world evaluation, we use three vision-language-action baselines: GR00T-N1.6~\citep{gr00tn1}, $\Psi_0$~\citep{wei2026psi0}, and $\pi_{0.5}$~\citep{intelligencepi05}.

All experiments use a multi-task training setting, including Being-M0.7, all baselines, and all ablation variants. In simulation, each method or variant is trained jointly on demonstrations from all six SIMPLE tasks, with a single model evaluated across the six tasks. For real-world experiments, each method is trained jointly on demonstrations from all three real-world tasks, with a single model evaluated across the three tasks. In both settings, the model is conditioned on the corresponding task instruction during evaluation to specify the task to be performed.

\subsection{Simulation Experiments}
\label{sec:simulation}

We evaluate Being-M0.7 on the SIMPLE benchmark~\citep{wei2026simple}, which provides a controlled environment for measuring whole-body locomotion and manipulation. We evaluate six tasks at three domain-randomization levels, from easy to hard, with ten trials per task and level: \textbf{BendPick}, \textbf{Handover}, \textbf{Mobile P\&P}, \textbf{Grasp}, \textbf{XMoveBendPick}, and \textbf{XMovePick}.

\paragraph{Main Comparison with Baselines.}
\begin{table}[ht]
\vspace{-12pt}
\caption{\textbf{Simulation results on SIMPLE.} Entries report successes out of ten trials at easy / medium / hard domain-randomization levels. Total aggregates six tasks and three levels (180 trials).}
\label{tab:benchmark}
\setlength{\tabcolsep}{3pt}
\begin{center}
\begin{tabular*}{\textwidth}{@{\extracolsep{\fill}}lccccccc@{}}
\toprule
\makecell[l]{\textbf{Method}}
& \makecell{\textbf{Bend}\\\textbf{Pick}}
& \makecell{\textbf{Handover}}
& \makecell{\textbf{Mobile}\\\textbf{P\&P}}
& \makecell{\textbf{Grasp}}
& \makecell{\textbf{XMove}\\\textbf{BendPick}}
& \makecell{\textbf{XMove}\\\textbf{Pick}}
& \makecell{\textbf{Total}} \\
\midrule
GR00T-N1.6 & 1/7/3 & 2/3/3 & 0/0/0 & 7/8/7 & 6/6/8 & 0/0/0 & 61/180 \\
$\Psi_0$ & 10/8/9 & 6/8/6 & 0/0/0 & 10/9/9 & 7/8/5 & 9/9/3 & 116/180 \\
$\pi_{0.5}$ & 0/0/0 & 5/7/7 & 7/6/6 & 7/8/5 & 10/9/7 & 6/2/3 & 95/180 \\
\midrule
\textbf{Being-M0.7 (Ours)} & 7/7/8 & 9/9/10 & 7/4/2 & 10/10/8 & 2/4/3 & 10/10/8 & \textbf{128/180} \\
\bottomrule
\end{tabular*}
\end{center}
\end{table}

Table~\ref{tab:benchmark} reports successes per setting. Being-M0.7 achieves the highest aggregate success rate, with $128/180$ successes ($71.1\%$), compared with $116/180$ ($64.4\%$) for $\Psi_0$, $95/180$ ($52.8\%$) for $\pi_{0.5}$, and $61/180$ ($33.9\%$) for GR00T-N1.6. Our method achieves $9/9/10$ on Handover and $10/10/8$ on both Grasp and XMovePick. On easy Mobile P\&P, it achieves $7/10$, matching $\pi_{0.5}$ and exceeding GR00T-N1.6 and $\Psi_0$, both at $0/10$. When results are aggregated across difficulty levels, $\Psi_0$ leads on BendPick, while $\pi_{0.5}$ leads on Mobile P\&P and XMoveBendPick.

\subsection{Ablation Study}
\label{sec:ablation}

\begin{table}[ht]
\caption{\textbf{Ablation results on SIMPLE.} Entries report successes out of ten trials at easy / medium / hard domain-randomization levels (180 trials total). Language-only context replaces prior predictions with task-language features; the expert retains current images and proprioception.}
\label{tab:simulation_ablation}
\setlength{\tabcolsep}{3pt}
\begin{center}
\begin{tabular*}{\textwidth}{@{\extracolsep{\fill}}lccccccc@{}}
\toprule
\makecell[l]{\textbf{Variant}}
& \makecell{\textbf{Bend}\\\textbf{Pick}}
& \makecell{\textbf{Handover}}
& \makecell{\textbf{Mobile}\\\textbf{P\&P}}
& \makecell{\textbf{Grasp}}
& \makecell{\textbf{XMove}\\\textbf{BendPick}}
& \makecell{\textbf{XMove}\\\textbf{Pick}}
& \makecell{\textbf{Total}} \\
\midrule
\textbf{Being-M0.7 (Full)} & 7/7/8 & 9/9/10 & 7/4/2 & 10/10/8 & 2/4/3 & 10/10/8 & \textbf{128/180} \\
\midrule
w/o pre-training & 0/0/0 & 9/10/9 & 2/2/3 & 8/8/5 & 5/2/4 & 8/6/7 & 88/180 \\
w/o mid-training & 2/0/0 & 9/9/8 & 7/3/4 & 8/9/7 & 5/6/4 & 9/7/10 & 107/180 \\
Language-only context & 4/4/5 & 8/9/8 & 4/4/2 & 9/10/9 & 7/5/9 & 4/2/2 & 105/180 \\
\bottomrule
\end{tabular*}
\end{center}
\vspace{-10pt}
\end{table}

We evaluate three variants on the same six SIMPLE tasks and three difficulty levels (180 trials per variant): \textbf{w/o pre-training} omits the prior model's mixed-modality pre-training stage; \textbf{w/o mid-training} omits its adaptation to robot trajectories; and \textbf{language-only context} replaces the upstream prior model's predictive context supplied to the action expert with language features from the task instruction. In the language-only variant, the expert still receives the current egocentric image and proprioceptive state $O_q$; only the upstream conditioning context is replaced.

\paragraph{Contribution of Training Stages.}
As shown in Table~\ref{tab:simulation_ablation}, removing pre-training reduces aggregate success from $128/180$ ($71.1\%$) to $88/180$ ($48.9\%$), a drop of $22.2$ percentage points. Removing robot mid-training yields $107/180$ ($59.4\%$), a drop of $11.7$ points. The largest task-level degradation in both variants occurs on BendPick, where successes fall from $22/30$ to $0/30$ and $2/30$, respectively. These aggregate results support the contributions of both video--motion prior learning during pre-training and subsequent robot-domain adaptation during mid-training.

\paragraph{Utility of Upstream Predictive Context.}
Replacing the prior model's output context with language-only features reduces aggregate success to $105/180$ ($58.3\%$), a drop of $12.8$ percentage points from the full model. The full model improves over this variant on four of the six tasks, with the largest gain on XMovePick ($28/30$ versus $8/30$). Language-only context performs better on XMoveBendPick ($21/30$ versus $9/30$), while both variants obtain $28/30$ on Grasp. Thus, the benefit varies across tasks, but the aggregate improvement indicates that the upstream prior model provides useful predictive information for downstream action generation beyond task language and current robot observations. We discuss possible interference among visually similar tasks in Appendix~\ref{app:simple_rollouts}.

\subsection{Real-World Experiments}

\begin{figure*}[ht]
    \centering
    \includegraphics[width=\linewidth]{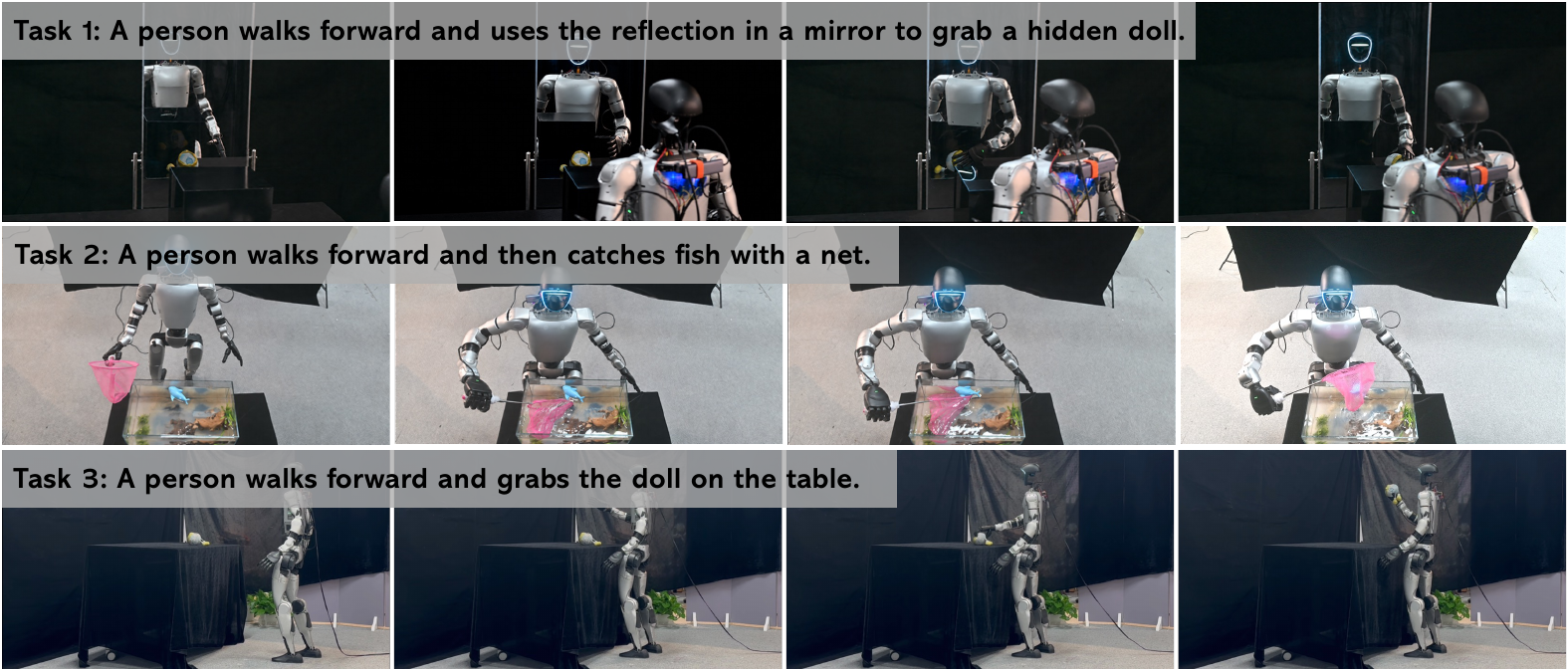}

    \caption{\textbf{Representative real-world execution sequences.} Rows show the three evaluated tasks, from top to bottom: \textbf{Walk to Mirror (W2M)}, approaching the table and grasping a hidden doll using mirror observations; \textbf{Walk to Fish (W2F)}, approaching the tank and scooping a toy fish with a handheld net; and \textbf{Walk to Doll (W2D)}, approaching the table and grasping a directly visible doll.}
    \label{fig:real_cases}
\end{figure*}

We quantitatively evaluate Being-M0.7 on three real-world loco-manipulation tasks: Walk to Fish \textbf{(W2F)}, Walk to Mirror \textbf{(W2M)}, and Walk to Doll \textbf{(W2D)}. All three tasks require the robot to approach the target before manipulating it, testing coordinated locomotion and manipulation alongside tool use, reflection-based reasoning, or direct grasping. We compare with GR00T-N1.6~\citep{gr00tn1}, $\Psi_0$~\citep{wei2026psi0}, and $\pi_{0.5}$~\citep{intelligencepi05}. Figure~\ref{fig:real_cases} illustrates these tasks.

\paragraph{Quantitative Evaluation.}

\begin{table}[ht]
\vspace{-12pt}
\caption{\textbf{Quantitative real-world evaluation.} Each task is evaluated over five trials. Entries report successful trials / total trials; bold numbers indicate the best result, including ties.}
\label{tab:real_world_all}
\begin{center}
\begin{tabular}{lcccc}
\toprule
Method & W2F & W2M & W2D & Overall \\
\midrule
GR00T-N1.6 & \textbf{3/5} & \textbf{5/5} & \textbf{5/5} & \textbf{13/15} \\
$\Psi_0$ & 0/5 & 1/5 & 2/5 & 3/15 \\
$\pi_{0.5}$ & 2/5 & 2/5 & 3/5 & 7/15 \\
\midrule
\textbf{Being-M0.7 (Ours)} & \textbf{3/5} & \textbf{5/5} & \textbf{5/5} & \textbf{13/15} \\
\bottomrule
\end{tabular}
\end{center}
\end{table}

We evaluate each method over five trials per task, totaling fifteen trials. Physical markers fix the robot starting positions and table positions across methods, while target objects are randomly initialized across the left, center, and right regions of the workspace. Each trial has a 30-second limit and is judged by a human evaluator: success requires grasping the doll in W2M and W2D, or scooping up the toy fish in W2F. As shown in Table~\ref{tab:real_world_all}, Being-M0.7 achieves $13/15$ successful trials ($86.7\%$), matching GR00T-N1.6 and exceeding $\pi_{0.5}$ ($7/15$) and $\Psi_0$ ($3/15$).

\begin{itemize}
\item \textbf{Walk to Fish.} The robot approaches the water tank and uses a handheld net to scoop up a toy fish. This evaluates the complete locomotion-and-tool-use sequence. Being-M0.7 and GR00T-N1.6 each succeed in $3/5$ trials, compared with $2/5$ for $\pi_{0.5}$ and $0/5$ for $\Psi_0$.

\item \textbf{Walk to Mirror.} The robot approaches a table, infers the location of a hidden toy from its mirror reflection, and grasps it. This task combines visual reasoning under partial observability with whole-body positioning and manipulation. Being-M0.7 and GR00T-N1.6 each succeed in $5/5$ trials, compared with $2/5$ for $\pi_{0.5}$ and $1/5$ for $\Psi_0$.

\item \textbf{Walk to Doll.} The robot approaches a table and directly grasps a visible doll. This task evaluates coordinated locomotion and target-directed grasping during the transition from approach to manipulation, without the mirror-based observation requirement. Being-M0.7 and GR00T-N1.6 each succeed in $5/5$ trials, compared with $3/5$ for $\pi_{0.5}$ and $2/5$ for $\Psi_0$.
\end{itemize}

\paragraph{Qualitative Case Studies.}
Figure~\ref{fig:real_cases} shows execution sequences for the three evaluated tasks. From top to bottom, W2M shows approaching the table and grasping a hidden doll using mirror observations; W2F shows approaching the water tank and scooping a toy fish with a net; and W2D shows approaching the table and grasping a directly visible doll. These sequences illustrate the transition from locomotion to manipulation in the corresponding task settings. Detailed task descriptions and additional snapshots are provided in Appendix~\ref{app:real_exp}.

\section{Conclusion}

We presented Being-M0.7, a latent world-action model that transfers visual--motion priors learned from mixed-modality human data to humanoid loco-manipulation. Joint latent visual and motion prediction learns a predictive prior, robot mid-training adapts it to robot trajectories, and a gated action expert combines the frozen prior's predictive context with current observations to generate executable whole-body commands for body control. Experiments on SIMPLE and real-world Unitree G1 tasks demonstrate the effectiveness of this framework, while ablations support the contributions of pre-training, robot mid-training, and predictive context to downstream control.

\paragraph{Limitations and Future Work.}
Our evaluation covers a limited range of task horizons, and performance on longer-horizon loco-manipulation tasks remains to be established. In our current setup, the SONIC whole-body controller also has limited end-effector tracking accuracy, which makes robot demonstration collection challenging. Future work will extend the framework to longer-horizon tasks and refine SONIC to improve tracking precision and facilitate demonstration collection.

\clearpage
\bibliographystyle{unsrt}
\bibliography{iclr2027_conference}

\begin{thebibliography}{10}

\bibitem{he2025omnih2o}
Tairan He, Zhengyi Luo, Xialin He, Wenli Xiao, Chong Zhang, Weinan Zhang,
  Kris~M Kitani, Changliu Liu, and Guanya Shi.
\newblock Omnih2o: Universal and dexterous human-to-humanoid whole-body
  teleoperation and learning.
\newblock In {\em Conference on Robot Learning}, pages 1516--1540. PMLR, 2025.

\bibitem{ben2025homie}
Qingwei Ben, Feiyu Jia, Jia Zeng, Junting Dong, Dahua Lin, and Jiangmiao Pang.
\newblock Homie: Humanoid loco-manipulation with isomorphic exoskeleton
  cockpit.
\newblock {\em arXiv preprint arXiv:2502.13013}, 2025.

\bibitem{fu2025humanplus}
Zipeng Fu, Qingqing Zhao, Qi~Wu, Gordon Wetzstein, and Chelsea Finn.
\newblock Humanplus: Humanoid shadowing and imitation from humans.
\newblock In {\em Conference on Robot Learning}, pages 2828--2844. PMLR, 2025.

\bibitem{wei2026psi0}
Songlin Wei, Hongyi Jing, Boqian Li, Zhenyu Zhao, Jiageng Mao, Zhenhao Ni,
  Sicheng He, Jie Liu, Xiawei Liu, Kaidi Kang, et~al.
\newblock {$\Psi_0$}: An open foundation model towards universal humanoid
  loco-manipulation.
\newblock {\em arXiv preprint arXiv:2603.12263}, 2026.

\bibitem{jiang2026wholebodyvla}
Haoran Jiang, Jin Chen, Qingwen Bu, Li~Chen, Modi Shi, Yanjie Zhang, Delong Li,
  Chuanzhe Suo, wang chuang, zhihui peng, and Hongyang Li.
\newblock Wholebody{VLA}: Towards unified latent {VLA} for whole-body
  loco-manipulation control.
\newblock In {\em The Fourteenth International Conference on Learning
  Representations}, 2026.

\bibitem{shi2026egohumanoid}
Modi Shi, Shijia Peng, Jin Chen, Haoran Jiang, Tianyu Li, Di~Huang, Ping Luo,
  Hongyang Li, and Li~Chen.
\newblock Egohumanoid: Unlocking in-the-wild loco-manipulation with robot-free
  egocentric demonstration.
\newblock {\em arXiv preprint arXiv:2602.10106}, 2026.

\bibitem{brohan2022rt1}
Anthony Brohan, Noah Brown, Justice Carbajal, Yevgen Chebotar, Joseph Dabis,
  Chelsea Finn, Keerthana Gopalakrishnan, Karol Hausman, Alex Herzog, Jasmine
  Hsu, et~al.
\newblock Rt-1: Robotics transformer for real-world control at scale.
\newblock {\em arXiv preprint arXiv:2212.06817}, 2022.

\bibitem{zitkovich2023rt2}
Brianna Zitkovich, Tianhe Yu, Sichun Xu, Peng Xu, Ted Xiao, Fei Xia, Jialin Wu,
  Paul Wohlhart, Stefan Welker, Ayzaan Wahid, et~al.
\newblock Rt-2: Vision-language-action models transfer web knowledge to robotic
  control.
\newblock In {\em Conference on Robot Learning}, pages 2165--2183. PMLR, 2023.

\bibitem{o2024open}
Abby O’Neill, Abdul Rehman, Abhiram Maddukuri, Abhishek Gupta, Abhishek
  Padalkar, Abraham Lee, Acorn Pooley, Agrim Gupta, Ajay Mandlekar, Ajinkya
  Jain, et~al.
\newblock Open x-embodiment: Robotic learning datasets and rt-x models: Open
  x-embodiment collaboration 0.
\newblock In {\em 2024 IEEE International Conference on Robotics and Automation
  (ICRA)}, pages 6892--6903. IEEE, 2024.

\bibitem{octo2024}
{Octo Model Team}, Dibya Ghosh, Homer Walke, Karl Pertsch, Kevin Black, Oier
  Mees, Sudeep Dasari, Joey Hejna, Tobias Kreiman, Charles Xu, et~al.
\newblock Octo: An open-source generalist robot policy.
\newblock {\em arXiv preprint arXiv:2405.12213}, 2024.

\bibitem{pi0}
Kevin Black, Noah Brown, Danny Driess, Adnan Esmail, Michael Equi, Chelsea
  Finn, Niccolo Fusai, Lachy Groom, Karol Hausman, Brian Ichter, et~al.
\newblock {$\pi_{0}$}: A vision-language-action flow model for general robot
  control.
\newblock {\em arXiv preprint arXiv:2410.24164}, 2024.

\bibitem{kim2025openvla}
Moo~Jin Kim, Karl Pertsch, Siddharth Karamcheti, Ted Xiao, Ashwin Balakrishna,
  Suraj Nair, Rafael Rafailov, Ethan~P Foster, Pannag~R Sanketi, Quan Vuong,
  et~al.
\newblock Openvla: An open-source vision-language-action model.
\newblock In {\em Conference on Robot Learning}, pages 2679--2713. PMLR, 2025.

\bibitem{du2023learning}
Yilun Du, Sherry Yang, Bo~Dai, Hanjun Dai, Ofir Nachum, Josh Tenenbaum, Dale
  Schuurmans, and Pieter Abbeel.
\newblock Learning universal policies via text-guided video generation.
\newblock {\em Advances in neural information processing systems},
  36:9156--9172, 2023.

\bibitem{agarwal2025cosmos}
Niket Agarwal, Arslan Ali, Maciej Bala, Yogesh Balaji, Erik Barker, Tiffany
  Cai, Prithvijit Chattopadhyay, Yongxin Chen, Yin Cui, Yifan Ding, et~al.
\newblock Cosmos world foundation model platform for physical ai.
\newblock {\em arXiv preprint arXiv:2501.03575}, 2025.

\bibitem{wam}
Seonghyeon Ye, Yunhao Ge, Kaiyuan Zheng, Shenyuan Gao, Sihyun Yu, George
  Kurian, Suneel Indupuru, You~Liang Tan, Chuning Zhu, Jiannan Xiang, et~al.
\newblock World action models are zero-shot policies.
\newblock {\em arXiv preprint arXiv:2602.15922}, 2026.

\bibitem{h05}
Hao Luo, Ye~Wang, Wanpeng Zhang, Sipeng Zheng, Ziheng Xi, Chaoyi Xu, Haiweng
  Xu, Haoqi Yuan, Chi Zhang, Yiqing Wang, et~al.
\newblock Being-h0. 5: Scaling human-centric robot learning for
  cross-embodiment generalization.
\newblock {\em arXiv preprint arXiv:2601.12993}, 2026.

\bibitem{h07}
Hao Luo, Wanpeng Zhang, Yicheng Feng, Sipeng Zheng, Haiweng Xu, Chaoyi Xu,
  Ziheng Xi, Yuhui Fu, and Zongqing Lu.
\newblock Being-h0. 7: A latent world-action model from egocentric videos.
\newblock {\em arXiv preprint arXiv:2605.00078}, 2026.

\bibitem{grauman2022ego4d}
Kristen Grauman, Andrew Westbury, Eugene Byrne, Zachary Chavis, Antonino
  Furnari, Rohit Girdhar, Jackson Hamburger, Hao Jiang, Miao Liu, Xingyu Liu,
  et~al.
\newblock Ego4d: Around the world in 3,000 hours of egocentric video.
\newblock In {\em Proceedings of the IEEE/CVF conference on computer vision and
  pattern recognition}, pages 18995--19012, 2022.

\bibitem{nymeria}
Lingni Ma, Yuting Ye, Fangzhou Hong, Vladimir Guzov, Yifeng Jiang, Rowan
  Postyeni, Luis Pesqueira, Alexander Gamino, Vijay Baiyya, Hyo~Jin Kim, et~al.
\newblock Nymeria: A massive collection of multimodal egocentric daily motion
  in the wild.
\newblock In {\em European Conference on Computer Vision}, pages 445--465.
  Springer, 2024.

\bibitem{guo2022generating}
Chuan Guo, Shihao Zou, Xinxin Zuo, Sen Wang, Wei Ji, Xingyu Li, and Li~Cheng.
\newblock Generating diverse and natural 3d human motions from text.
\newblock In {\em Proceedings of the IEEE/CVF Conference on Computer Vision and
  Pattern Recognition}, pages 5152--5161, 2022.

\bibitem{snap}
chuan guo, Inwoo Hwang, Jian Wang, and Bing Zhou.
\newblock Snapmogen: Human motion generation from expressive texts.
\newblock In {\em Advances in Neural Information Processing Systems},
  volume~38, pages 99939--99955, 2025.

\bibitem{zhao2025resmimic}
Siheng Zhao, Yanjie Ze, Yue Wang, C~Karen Liu, Pieter Abbeel, Guanya Shi, and
  Rocky Duan.
\newblock Resmimic: From general motion tracking to humanoid whole-body
  loco-manipulation via residual learning.
\newblock {\em arXiv preprint arXiv:2510.05070}, 2025.

\bibitem{weng2025hdmi}
Haoyang Weng, Yitang Li, Nikhil Sobanbabu, Zihan Wang, Zhengyi Luo, Tairan He,
  Deva Ramanan, and Guanya Shi.
\newblock Hdmi: Learning interactive humanoid whole-body control from human
  videos.
\newblock {\em arXiv preprint arXiv:2509.16757}, 2025.

\bibitem{lessmimic}
Yutang Lin, Jieming Cui, Yixuan Li, Baoxiong Jia, Yixin Zhu, and Siyuan Huang.
\newblock Lessmimic: Long-horizon humanoid interaction with unified distance
  field representations.
\newblock {\em arXiv preprint arXiv:2602.21723}, 2026.

\bibitem{xu2025interact}
Sirui Xu, Dongting Li, Yucheng Zhang, Xiyan Xu, Qi~Long, Ziyin Wang, Yunzhi Lu,
  Shuchang Dong, Hezi Jiang, Akshat Gupta, Yu-Xiong Wang, and Liang-Yan Gui.
\newblock {InterAct}: Advancing large-scale versatile 3d human-object
  interaction generation.
\newblock In {\em CVPR}, 2025.

\bibitem{yin2025visualmimic}
Shaofeng Yin, Yanjie Ze, Hong-Xing Yu, C~Karen Liu, and Jiajun Wu.
\newblock Visualmimic: Visual humanoid loco-manipulation via motion tracking
  and generation.
\newblock {\em arXiv preprint arXiv:2509.20322}, 2025.

\bibitem{tevet2022human}
Guy Tevet, Sigal Raab, Brian Gordon, Yonatan Shafir, Daniel Cohen-Or, and
  Amit~H. Bermano.
\newblock Human motion diffusion model.
\newblock {\em arXiv preprint arXiv:2209.14916}, 2022.

\bibitem{jiang2023motiongpt}
Biao Jiang, Xin Chen, Wen Liu, Jingyi Yu, Gang Yu, and Tao Chen.
\newblock Motiongpt: Human motion as a foreign language.
\newblock {\em Advances in Neural Information Processing Systems},
  36:20067--20079, 2023.

\bibitem{li2023egoego}
Jiaman Li, C.~Karen Liu, and Jiajun Wu.
\newblock Ego-body pose estimation via ego-head pose estimation.
\newblock In {\em Proceedings of the IEEE/CVF Conference on Computer Vision and
  Pattern Recognition}, 2023.

\bibitem{yi2025egoallo}
Brent Yi, Vickie Ye, Maya Zheng, Yunqi Li, Lea M{\"u}ller, Georgios Pavlakos,
  Yi~Ma, Jitendra Malik, and Angjoo Kanazawa.
\newblock Estimating body and hand motion in an ego-sensed world.
\newblock In {\em Proceedings of the IEEE/CVF Conference on Computer Vision and
  Pattern Recognition}, 2025.

\bibitem{patel2025uniegomotion}
Chaitanya Patel, Hiroki Nakamura, Yuta Kyuragi, Kazuki Kozuka, Juan~Carlos
  Niebles, and Ehsan Adeli.
\newblock Uniegomotion: A unified model for egocentric motion reconstruction,
  forecasting, and generation.
\newblock In {\em Proceedings of the IEEE/CVF International Conference on
  Computer Vision}, 2025.

\bibitem{chen2025egoagent}
Lu~Chen, Yizhou Wang, Shixiang Tang, Qianhong Ma, Tong He, Wanli Ouyang,
  Xiaowei Zhou, Hujun Bao, and Sida Peng.
\newblock Acquisition through my eyes and steps: A joint predictive agent model
  in egocentric worlds.
\newblock In {\em Proceedings of the IEEE/CVF International Conference on
  Computer Vision}, 2025.

\bibitem{xiu2025egotwin}
Jingqiao Xiu, Fangzhou Hong, Yicong Li, Mengze Li, Wentao Wang, Sirui Han,
  Liang Pan, and Ziwei Liu.
\newblock Egotwin: Dreaming body and view in first person.
\newblock {\em arXiv preprint arXiv:2508.13013}, 2025.

\bibitem{gr00tn1}
Johan Bjorck, Fernando Casta{\~n}eda, Nikita Cherniadev, Xingye Da, Runyu Ding,
  Linxi Fan, Yu~Fang, Dieter Fox, Fengyuan Hu, Spencer Huang, et~al.
\newblock Gr00t n1: An open foundation model for generalist humanoid robots.
\newblock {\em arXiv preprint arXiv:2503.14734}, 2025.

\bibitem{zheng2026motionwam}
Jia Zheng, Teli Ma, Yudong Fan, Zifan Wang, Shuo Yang, and Junwei Liang.
\newblock Motionwam: Towards foundation world action models for real-time
  humanoid loco-manipulation.
\newblock {\em arXiv preprint arXiv:2606.09215}, 2026.

\bibitem{simeoni2025dinov3}
Oriane Sim{\'e}oni, Huy~V. Vo, Maximilian Seitzer, Federico Baldassarre, Maxime
  Oquab, Cijo Jose, Vasil Khalidov, Marc Szafraniec, Seungeun Yi, Micha{\"e}l
  Ramamonjisoa, et~al.
\newblock Dinov3.
\newblock {\em arXiv preprint arXiv:2508.10104}, 2025.

\bibitem{xie2023future}
Zhihui Xie, Zichuan Lin, Deheng Ye, Qiang Fu, Yang Wei, and Shuai Li.
\newblock Future-conditioned unsupervised pretraining for decision transformer.
\newblock In {\em International Conference on Machine Learning}, pages
  38187--38203. PMLR, 2023.

\bibitem{xperience}
{Ropedia}.
\newblock {Xperience-10M}: A large-scale egocentric multimodal dataset with
  structured {3D}/{4D} annotations.
\newblock Hugging Face dataset, 2026.
\newblock \url{https://huggingface.co/datasets/ropedia-ai/xperience-10m},
  accessed July 11, 2026.

\bibitem{luo2025sonic}
Zhengyi Luo, Ye~Yuan, Tingwu Wang, Chenran Li, Fernando Casta{\~n}eda, Sirui
  Chen, Zi-Ang Cao, Jiefeng Li, David Minor, Qingwei Ben, et~al.
\newblock Sonic: Supersizing motion tracking for natural humanoid whole-body
  control.
\newblock {\em arXiv preprint arXiv:2511.07820}, 2025.

\bibitem{lafan1}
F{\'e}lix~G Harvey, Mike Yurick, Derek Nowrouzezahrai, and Christopher Pal.
\newblock Robust motion in-betweening.
\newblock {\em ACM Transactions on Graphics (TOG)}, 39(4):60--1, 2020.

\bibitem{m05}
Bin Cao, Sipeng Zheng, Ye~Wang, Lujie Xia, Qianshan Wei, Qin Jin, Jing Liu, and
  Zongqing Lu.
\newblock Being-m0. 5: A real-time controllable vision-language-motion model.
\newblock {\em arXiv preprint arXiv:2508.07863}, 2025.

\bibitem{opent2m}
Bin Cao, Sipeng Zheng, Hao Luo, Boyuan Li, Jing Liu, and Zongqing Lu.
\newblock Opent2m: No-frill motion generation with open-source, large-scale,
  high-quality data.
\newblock In {\em Proceedings of the IEEE/CVF Conference on Computer Vision and
  Pattern Recognition}, pages 30640--30649, 2026.

\bibitem{liang2024mixtureoftransformers}
Weixin Liang, Lili Yu, Liang Luo, Srinivasan Iyer, Ning Dong, Chunting Zhou,
  Gargi Ghosh, Mike Lewis, Wen-tau Yih, Luke Zettlemoyer, and Xi~Victoria Lin.
\newblock Mixture-of-transformers: A sparse and scalable architecture for
  multi-modal foundation models.
\newblock {\em arXiv preprint arXiv:2411.04996}, 2024.

\bibitem{sanh2019distilbert}
Victor Sanh, Lysandre Debut, Julien Chaumond, and Thomas Wolf.
\newblock Distilbert, a distilled version of bert: smaller, faster, cheaper and
  lighter.
\newblock {\em arXiv preprint arXiv:1910.01108}, 2019.

\bibitem{lipman2023flowmatching}
Yaron Lipman, Ricky T.~Q. Chen, Heli Ben-Hamu, Maximilian Nickel, and Matt Le.
\newblock Flow matching for generative modeling.
\newblock In {\em International Conference on Learning Representations}, 2023.

\bibitem{zhao2023learning}
Tony~Z. Zhao, Vikash Kumar, Sergey Levine, and Chelsea Finn.
\newblock Learning fine-grained bimanual manipulation with low-cost hardware.
\newblock In {\em Proceedings of Robotics: Science and Systems}, Daegu,
  Republic of Korea, July 2023.

\bibitem{zhao2026xrobotoolkit}
Zhigen Zhao, Liuchuan Yu, Ke~Jing, and Ning Yang.
\newblock Xrobotoolkit: A cross-platform framework for robot teleoperation.
\newblock In {\em 2026 IEEE/SICE International Symposium on System Integration
  (SII)}, pages 15--20. IEEE, 2026.

\bibitem{intelligencepi05}
Physical Intelligence, Kevin Black, Noah Brown, James Darpinian, Karan
  Dhabalia, Danny Driess, Adnan Esmail, Michael Equi, Chelsea Finn, Niccolo
  Fusai, et~al.
\newblock {$\pi_{0.5}$}: A vision-language-action model with open-world
  generalization.
\newblock {\em arXiv preprint arXiv:2504.16054}, 2025.

\bibitem{wei2026simple}
Songlin Wei, Zhenhao Ni, Jie Liu, Zhenyu Zhao, Junjie Ye, Hongyi Jing, Junkai
  Xia, Xiawei Liu, Michael Leong, Liang Heng, et~al.
\newblock Simple: Simulation-based policy learning and evaluation for humanoid
  loco-manipulation.
\newblock {\em arXiv preprint arXiv:2606.08278}, 2026.

\bibitem{peebles2023dit}
William Peebles and Saining Xie.
\newblock Scalable diffusion models with transformers.
\newblock In {\em Proceedings of the IEEE/CVF International Conference on
  Computer Vision}, 2023.

\end{thebibliography}

\clearpage
\beginappendix
\section*{Author List}
\paragraph{Core Contributors: } Junpeng Yue$^*$, Boyuan Li$^*$, Yuxuan Wang$^*$, Zepeng Wang, Yuhui Fu, Feiyang Xie, Yu Zhang, Jing Zhang,  Jiangxing Wang, Zongqing Lu$^{\dagger}$ \footnote{$^*$Equal Contribution \quad $^{\dagger}$Corresponding Author}

\paragraph{Contributors: } Weibo Li, Xiaofei Zheng, Yuming Fang

\section{Additional Method Details}
\subsection{Data Sources and Motion Processing}
\label{app:data_details}
Figure~\ref{fig:data} summarizes the raw pre-training corpus and its three supervision streams. Source hours refer to the collection before filtering and temporal segmentation.

\paragraph{Head-root motion vector.}
The shared 22D representation uses the head as the root for both human and humanoid motion. In storage order, the vector contains root translation (dimensions 1--3), root rotation in a continuous 6D representation (4--9), initial head height $h_0$ (10), and the local 3D positions of the left hand (11--13), right hand (14--16), left foot (17--19), and right foot (20--22). The 6D rotation encodes two axes of the head orientation; the decoder orthonormalizes them and obtains the third axis by a cross product. Positions and heights are measured in meters, and rotation coordinates are dimensionless.

Let $p_t$ and $R_t$ be the global head position and orientation, and $p_t^e$ the global position of end-effector $e$. The canonical frame is floor-aligned with $y$ as the vertical axis. Its origin lies below the initial head position, so $p_0=(0,h_0,0)$. Human motion is expressed in this head-root frame by retaining the head pose and transforming the two hand and two foot positions as
\[
\ell_t^e=R_t^\top(p_t^e-p_t),\qquad
p_t^e=p_t+R_t\ell_t^e.
\]
Robot motion uses the same component order and coordinate convention; the robot data loader reads the preprocessed 22D trajectories stored alongside each recorded episode. The shared representation therefore describes head motion and end-effector positions without requiring identical human and robot joint-angle parameterizations.

\paragraph{Translation encoding and resampling.}
The stored translation channels contain frame-to-frame head displacements $\delta p_t=(\delta x_t,\delta y_t,\delta z_t)$ in the canonical global frame. The initial displacement is reset to zero when constructing a robot window. With \texttt{use\_root\_pos=true}, these displacements are accumulated before subsampling to 5 Hz, so the model input contains relative head positions $d_t$ rather than per-frame displacements:
\[
d_t=\sum_{s\leq t}\delta p_s,\qquad
p_t=d_t+(0,h_0,0).
\]
The rotation, initial-height, and local-position channels retain their meanings during resampling. Thus, the first ten model-input dimensions are $[d_t;\operatorname{rot6D}(R_t);h_0]$, followed by the twelve local-position dimensions.

\paragraph{Normalization.}
Human-motion loaders apply per-channel standardization $(m-\mu)/\sigma$ when the dataset's stored mean and standard deviation are available; standard deviations below $10^{-5}$ are replaced by one. If these statistics are absent, the loader retains the unstandardized motion values. The robot mid-training loader uses unstandardized 22D motion vectors. This motion preprocessing is separate from post-training normalization of proprioception and executable commands, whose statistics are computed from the robot training split.

\subsection{Architecture Details}
\label{app:architecture_details}
\paragraph{Input Embeddings.}
We map per-frame visual latents $z_t$ and motion vectors $m_t$ from~\Cref{sec:problem-formulation} into a common MoT hidden space using separate learned projections. Appendix~\ref{app:data_details} details the motion representation. Temporal positional embeddings encode their aligned time indices, while modality-type embeddings distinguish visual and motion inputs. For video-only or motion-only samples, supervision is applied only to the available modality. The task instruction $I$ is encoded separately by the frozen DistilBERT~\citep{sanh2019distilbert} and supplied as language-conditioning features.

\paragraph{Missing Modalities.}
The model keeps a fixed interleaved layout of visual and motion token slots. For a video-only sample, projected motion features are replaced by a learned missing-motion embedding; for a motion-only sample, projected visual features are replaced by a learned missing-image embedding. Modality-presence flags are combined with temporal-validity masks to define the valid history and future tokens. Valid queries cannot attend to absent-modality keys or values. An invalid query is allowed to attend only to itself in this shared attention operation, preventing an all-masked attention row; its output is excluded from supervision and from the context supplied to the action expert. Hence, placeholder tokens do not exchange information with valid visual or motion tokens, while paired samples retain cross-modal attention.

During action post-training and deployment, \texttt{use\_robot\_motion=false} sets all motion-presence flags to false. The motion state is initialized to zero and receives no flow updates; only visual future states are integrated and cached for the expert. The implementation still evaluates the shared MoT forward pass, including the masked motion slots and motion output projection, whose predictions are discarded. Disabling motion prediction therefore removes motion conditioning, motion-state generation, and motion supervision, without pruning all motion-branch computation.

\paragraph{MoT Blocks.}
The prior model is implemented as a Mixture-of-Transformers (MoT)~\citep{liang2024mixtureoftransformers} with modality-specific computation and shared multimodal attention. In each block, latent visual and motion tokens use separate layer normalizations, query/key/value and output projections, and feed-forward networks. Their projected queries, keys, and values are then combined in a shared attention operation, allowing every valid latent visual token to exchange information with motion tokens across time while retaining modality-specific parameterization. Both streams additionally cross-attend to the instruction features.
A sinusoidal embedding of the flow timestep $\tau$ modulates each modality's attention, language cross-attention, and feed-forward computation through modality-specific adaptive layer normalization~\citep{peebles2023dit}.

The first $K$ video-motion frames are clean context tokens, while tokens in the future interval are corrupted at the sampled flow timestep. We use a chunk attention mask under which context tokens attend within the observed history but cannot access the future, whereas all noisy future tokens jointly attend to the full history and to one another. This non-causal target interaction lets the model denoise the entire future video-motion chunk in parallel. Finally, modality-specific adaptive normalization and output projections produce velocity fields in the visual latent space and the continuous motion space, respectively.

\begin{figure}[htbp]
    \centering
    \includegraphics[width=0.62\linewidth]{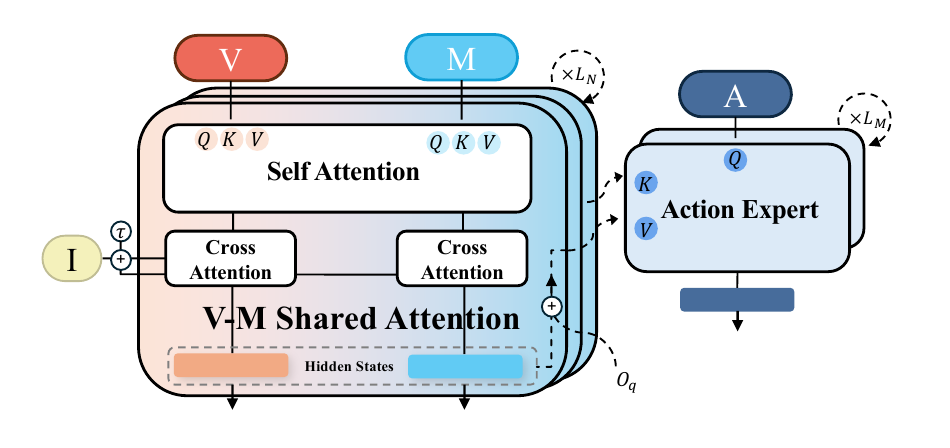}
    \caption{\textbf{Attention interfaces.} $V$, $M$, $I$, and $A$ denote vision, motion, instruction, and action hidden states; $\tau$ is the flow timestep. The prior model learns from visual and motion streams during pre-training and mid-training. During action post-training and inference, only its visual hidden states are supplied to the expert, alongside current robot observations $O_q$. The gated residual is defined in Eq.~\ref{eq:world_gate}.}
    \label{fig:attention_details}
\end{figure}

\paragraph{Layer-Wise Action Interface.}
At an expert layer connected to the prior model, local self-attention first processes the action and current-observation tokens. The subsequent world cross-attention uses only the noisy action tokens as queries; proprioceptive tokens do not query the prior model or receive this cross-attention residual. Proprioception still conditions action generation through local self-attention. Keys and values are constructed from the generated future visual-token hidden states before the corresponding layer of the prior model, excluding the prior model's observed-history input tokens. The projected cross-attention residual is applied only to the action tokens and gated as in Eq.~\ref{eq:world_gate}. This one-way connection allows the expert to access multiple representation depths without sending action information into the frozen prior model. Selected prior model layers and model dimensions are reported in Appendix~\ref{app:architecture_hyperparameters}.

\section{Implementation Details}
\subsection{Architecture Hyperparameters}
\label{app:architecture_hyperparameters}

Table~\ref{tab:architecture_hyperparameters} reports the model sizes used in our implementation. Layer indices of the prior model are one-based.

\begin{table}[ht]
\caption{Architecture hyperparameters of the video--motion prior model and action expert.}
\label{tab:architecture_hyperparameters}
\begin{center}
\begin{tabular}{l c}
\hline
\textbf{Hyperparameter} & \textbf{Value} \\
\hline
\multicolumn{2}{l}{\textbf{Video--Motion Prior Model}} \\
\hline
Transformer layers & 30 \\
Hidden width & 768 \\
Attention heads & 24 \\
Visual FFN expansion ratio & 6 \\
Motion FFN expansion ratio & 2 \\
Dropout & 0.1 \\
\hline
\multicolumn{2}{l}{\textbf{Action Expert}} \\
\hline
Transformer layers & 30 \\
Hidden width & 1024 \\
Attention heads & 32 \\
FFN expansion ratio & 4 \\
Connected prior model layers & $3,6,9,\ldots,30$ \\
Action chunk length & 30 \\
Dropout & 0.1 \\
\hline
\end{tabular}
\end{center}
\end{table}

\subsection{Hyperparameters}
\label{app:hyperparameters}

Table~\ref{tab:hyperparameters} reports the temporal and optimization settings. Effective batch sizes include all eight GPUs, with one optimizer update per batch and no gradient accumulation. The robot-stage settings follow the real-world mid-training and post-training configurations.

\begin{table}[!htb]
\centering
\caption{Temporal and training settings. The action-query interval is determined by executing a complete 30-command chunk at 50 Hz.}
\label{tab:hyperparameters}
\begin{tabular}{l c}
\toprule
\textbf{Temporal setting} & \textbf{Value} \\
\midrule
Sequence length $T$ / observed frames $K$ & 25 / 1 \\
Generated future frames / horizon & 24 / 4.8 s \\
Input resolution & $224\times224$ \\
Prior prediction frame rate & 5 Hz \\
Action chunk length $H_a$ & 30 \\
Action-query interval / rate & 0.6 s / $\approx1.67$ Hz \\
Command-execution interval / rate & 0.02 s / 50 Hz \\
Cache refresh & Before the prediction window is exhausted \\
Prior sampling steps during post-training & 3 \\
Prior / action sampling steps at inference & 10 / 10 \\
\bottomrule
\end{tabular}

\medskip
\begin{tabular}{l c c c}
\toprule
\textbf{Optimization setting} & \textbf{Pre-training} & \textbf{Mid-training} & \textbf{Post-training} \\
\midrule
GPUs & 8 & 8 & 8 \\
Batch size per GPU & 256 & 64 & 64 \\
Effective batch size & 2048 & 512 & 512 \\
Learning rate & $2\times10^{-4}$ & $10^{-4}$ & $10^{-4}$ \\
Optimizer & AdamW & AdamW & AdamW \\
Weight decay & 0.05 & 0.05 & 0.05 \\
Gradient norm clipping & 1.0 & 1.0 & 1.0 \\
Precision & FP16 mixed & FP16 mixed & FP16 mixed \\
Trainable module & Prior & Prior & Action expert \\
$\lambda_{\mathrm{local\text{-}vel}}$ & 1.0 & 0 & --- \\
World-context dropout & --- & --- & 0.1 \\
\bottomrule
\end{tabular}
\end{table}

\paragraph{Training data and sampling.}
Pre-training combines paired video--motion, video-only, and motion-only samples. The mixed-data loader shuffles the concatenated datasets without explicit source reweighting (\texttt{dataset\_weights=null}); therefore, the stream proportions are determined by the numbers of valid examples after filtering, rather than a fixed equal-modality mixture. Both robot stages use the same training and held-out episode lists. Mid-training reads video, text, and 22D motion, whereas post-training reads video, text, current proprioception, and demonstrated action chunks.

\paragraph{Optimization and initialization.}
The DINO and DistilBERT encoders remain frozen in all stages. Pre-training and mid-training update the prior model; mid-training starts from the last saved pre-training checkpoint with a fresh optimizer and learning-rate scheduler. The post-training configuration initializes the prior from a mid-training checkpoint at 50,000 optimizer updates, freezes it in evaluation mode, and trains only the action expert and its observation/context projections. All stages use linear learning-rate warmup followed by cosine decay. Pre-training checkpointing monitors validation local end-effector position error and retains the two best checkpoints and the latest checkpoint. Robot-stage checkpointing records validation loss and retains all saved checkpoints; the post-training metric is the command flow-matching loss.

\paragraph{Visual and language features.}
For the action expert's current image, we instead use the patch tokens from the final hidden layer, excluding the classification and register tokens. Images are resized to $224\times224$ and normalized with ImageNet channel statistics. The frozen \texttt{distilbert-base-uncased} encoder supplies final-layer token features, with instructions truncated to at most 64 tokens and padding masked in language cross-attention.

\subsection{Geometry-aware Auxiliary Losses}
\label{app:geometry_losses}

During pre-training, the motion branch uses two geometry-aware auxiliary losses in addition to the flow matching loss:
\[
\mathcal L_{\mathrm{geom}}
=\lambda_{\mathrm{traj}}\mathcal L_{\mathrm{traj}}
+\lambda_{\mathrm{local\text{-}vel}}\mathcal L_{\mathrm{local\text{-}vel}}.
\]
The clean-motion estimate can be recovered from the predicted velocity as
$$
\hat m_{0,t}=m_{\tau,t}-\tau u^M_\theta(m_{\tau,t},\tau,c,\mathcal{C}).
$$
Let $g(m_t)\in\mathbb{R}^3$ denote the decoded global head/root position, and let $\ell(m_t)$ denote the local end-effector coordinates. Both are decoded after undoing any motion normalization. Let $b_t$ indicate a valid, unpadded frame and $r^M$ indicate that motion is present, and define $\Omega_M=\{t\in\mathcal Q\mid b_t=1,\ r^M=1\}$.

The global trajectory auxiliary loss is
$$
\mathcal{L}_{\mathrm{traj}}
=\frac{1}{|\Omega_M|}
\sum_{t\in\Omega_M}
\left\|g(\hat m_{0,t})-g(m_t)\right\|_2.
$$

The local velocity auxiliary loss constrains frame-to-frame local end-effector motion. Define
$$
\Omega_M^\Delta
=\{t\mid t\in\mathcal{Q},\ b_t=1,\ b_{t-1}=1,\ r^M=1\}.
$$
Then
$$
\mathcal{L}_{\mathrm{local\text{-}vel}}
=\frac{1}{|\Omega_M^\Delta|d_\ell}
\sum_{t\in\Omega_M^\Delta}
\left\|
\Delta\ell(\hat m_{0,t})-\Delta\ell(m_t)
\right\|_2^2,
$$
where $\Delta\ell(m_t)=\ell(m_t)-\ell(m_{t-1})$ and $d_\ell=12$ for the two hands and two feet in the head/root frame. An auxiliary term is zero when its valid index set is empty. We use $(\lambda_{\mathrm{traj}},\lambda_{\mathrm{local\text{-}vel}})=(0.1,1.0)$ in pre-training.

\subsection{Post-training Details}

\label{app:posttrain}

\paragraph{Sampling from the Frozen Prior Model.}
During post-training, the observed visual history is clamped and explicit motion prediction is disabled. The frozen prior model initializes future visual latents from Gaussian noise and integrates its flow ODE using $N_{\mathrm{train}}=3$ sampling steps to generate a visual future. The expert's conditioning states are extracted along this sampled trajectory, rather than obtained by adding noise to ground-truth future latents. At each sampled world denoising timestep, we retain hidden states from the selected prior model layers. For an action flow timestep $\tau$, $H^\theta(\tau)$ uses the nearest sampled world denoising timestep. This procedure trains the expert on the prior model's own predictions, matching the source of predictive context used at inference. The three-step setting applies to sampling from the prior model during post-training; inference uses the ten-step schedule in Appendix~\ref{app:inference}. Gaussian corruption of demonstrated action chunks is still used for the expert's flow-matching objective.

\paragraph{Predictive Context Selection.}
Hidden states from all 24 generated future frames, at indices $\{K+1,\ldots,T\}$ with $K=1$ and $T=25$, are exposed to the action expert as world cross-attention keys and values. The observed input frame conditions the prior model internally but is excluded from the expert's cache. The expert receives its current robot observation separately through $O_q$. The same future-token selection is used during post-training and inference.

\paragraph{World-Context Dropout.}
For real-world experiments, we apply sample-level world-context dropout with probability 0.1 during post-training, with the same dropout decision shared across all expert layers. This encourages the action expert to use the current observation $O_q$ and limits over-reliance on the prior model's predictive context. World-context dropout is not used in simulation experiments.

\paragraph{Action Space.}

For real-world whole-body control, we use SONIC~\citep{luo2025sonic}. Each action is a 66D vector: a 64D SONIC universal control token followed by two scalar open/close commands, ordered as left hand and right hand. Recorded hand commands are binary values in $[0,1]$. Proprioceptive inputs and action targets are standardized per channel using the robot training-split statistics; predicted commands are transformed back to controller units before execution.

\paragraph{Robot Observation.}

The robot observation $O_q$ consists of the current egocentric image and a 63D proprioceptive vector containing the recorded roll and pitch, angular velocity, joint positions, and joint velocities. The image is encoded into patch tokens using the same frozen DINO encoder used for the visual stream. The post-training configuration disables the optional normalized execution-progress token (\texttt{policy\_use\_progress\_token=false}); the runtime command counter tracks progress within the prediction window but is not supplied as a separate policy token.

\paragraph{Prior Model Layer Selection.}

The prior model and action expert each have 30 layers, with corresponding layers paired by depth. World cross-attention is applied at every third expert layer, at one-based indices $3,6,9,\ldots,30$, using hidden states before the matching prior layer. The other expert layers process local observation and action tokens without injecting world context.

\paragraph{Temporal Configuration.}
Each prior window contains one observed frame ($K=1$) and 24 future frames ($T=25$). At a prediction frame rate of 5 Hz, the future spans $24/5=4.8$ seconds. All 24 future-token hidden states are retained as predictive context. At every action query, the expert receives the latest single image and proprioceptive state from the robot observation stream.

The three temporal rates have distinct roles. The prior's 5 Hz frame rate specifies the spacing of predicted visual states ($0.2$ seconds). The low-level controller executes commands at 50 Hz ($0.02$ seconds per command). Each expert query predicts an action chunk of length $H_a=30$, covering $30/50=0.6$ seconds, and a new query is made after that chunk is executed. The nominal query rate is therefore $1/0.6\approx1.67$ Hz. The prior context is refreshed before the prediction window is exhausted.

\subsection{Inference Details}
\label{app:inference}

Both the prior model and the action expert use $N=10$ Flow Matching steps with denoising timesteps
\[
\{\tau_n\}_{n=1}^{N}.
\]
During denoising by the prior model, we extract the intermediate representations
\[
\left\{
H_l^\theta(\tau_n)
\right\}_{l}
\]
from the selected prior model layers, retaining all 24 generated future positions specified in Appendix~\ref{app:posttrain}. These representations are converted into policy-side key-value caches; observed-history input tokens are not cached for the expert. Following the notation in the main text, we collectively denote the cached prior model context associated with denoising step $n$ as
\[
H_n^\theta.
\]
Because the prior model and the action expert use the same denoising schedule, $H_n^\theta$ is reused at the corresponding action denoising timestep $\tau_n$.

The action expert initializes the action chunk from Gaussian noise,
\[
\mathbf{a}^{0}\sim\mathcal{N}(0,I),
\]
and performs $N$ denoising steps. At step $n$, it predicts
\[
\hat{u}^{n}
=
u_{\phi}^{a}
\left(
\mathbf{a}^{n-1},
\tau_n,
H_n^\theta,
O_{\mathrm{cur}}
\right),
\]
where $O_{\mathrm{cur}}$ denotes the current robot observation. The action sample is updated as
\[
\mathbf{a}^{n}
=
\mathbf{a}^{n-1}
-
\hat{u}^{n}\Delta\tau_n,
\qquad
n=1,\ldots,N.
\]
The final output is the executable action chunk
\[
\mathbf{a}^{N}
=
(a_t,\ldots,a_{t+H_a-1}),
\qquad
H_a=30.
\]

At the start of a window, we acquire the conditioning image, generate the 24-frame future, and cache its hidden states. During execution, a command counter records the position within the current prediction window. Each action query uses the latest $O_{\mathrm{cur}}$ and the same cached future to produce 30 commands, which the low-level controller executes at 50 Hz. The next query occurs after $0.6$ seconds of command execution. Before the current prediction window is exhausted, a new image is acquired, the prior regenerates the future, the cache is replaced, and the counter is reset. Between cache refreshes, every action query remains conditioned on the latest robot observation. These rates describe the nominal execution schedule; denoising compute time is distinct from the duration represented by a command chunk.

\section{SIMPLE Benchmark and Simulation Task Details}
\label{app:sim_exp}

\subsection{Benchmark and Evaluation Protocol}
SIMPLE (SIMulation-based Policy Learning and Evaluation) is a simulation framework for humanoid loco-manipulation~\citep{wei2026simple}. It combines MuJoCo physics with Isaac Sim rendering and supports demonstration collection through motion planning and VR teleoperation. We evaluate six Unitree G1 tasks spanning tabletop grasping, bimanual transfer, and mobile manipulation. The \texttt{MP} and \texttt{Teleop} suffixes in the environment names identify the demonstration-generation pipelines.

We follow the \href{https://github.com/physical-superintelligence-lab/SIMPLE\#-simulation-benchmarking-results}{officially released six-task evaluation protocol}, retaining the official task implementations, initialization and domain-randomization configurations, and task-specific success evaluators. We use the default random-seed settings provided by the official implementation without overriding them. Each method is evaluated on ten trials for every task and randomization level, giving $6\times3\times10=180$ trials per method. The easy, medium, and hard settings correspond to Levels~0, 1, and~2, respectively. Table~\ref{tab:simple_levels} summarizes the cumulative perturbations. Higher levels add variation; individual task success counts need not decrease monotonically across levels.

All simulation baseline results in Table~\ref{tab:benchmark} are obtained from our own training and evaluation runs using the respective official implementations. All methods are evaluated on the same machine. The ablation variants in Table~\ref{tab:simulation_ablation} follow the same task configurations, default seed settings, success criteria, and trial counts as the main comparison. We report successful trials for each task and level, and compute the aggregate success rate over all 180 trials.

\begin{table}[!htb]
\centering
\caption{\textbf{SIMPLE evaluation levels.} Each level retains the variations of the preceding level. The definitions follow the official evaluation protocol.}
\label{tab:simple_levels}
\renewcommand{\arraystretch}{1.15}
\begin{tabularx}{\linewidth}{@{}l l X@{}}
\toprule
\textbf{Level} & \textbf{Table notation} & \textbf{Randomization} \\
\midrule
0 & Easy & Table materials; distractor identities and initial positions. \\
1 & Medium & Level 0 variations, plus substantial lighting changes. \\
2 & Hard & Level 1 variations, plus target-object initial-position perturbations. \\
\bottomrule
\end{tabularx}
\end{table}

\subsection{The Six Evaluated Tasks}
\label{app:simple_tasks}
Table~\ref{tab:simple_tasks} maps the task names in the main results to their simulation environments and summarizes their objectives~\citep{wei2026simple}. Together, the tasks vary the required reaching height, use of two hands, and integration of locomotion with object interaction.

\begin{table}[!htb]
\centering
\caption{\textbf{Task definitions and environment mapping.} Append each suffix below to \texttt{G1Wholebody} and prepend \texttt{simple/} to obtain the full environment identifier.}
\label{tab:simple_tasks}
\renewcommand{\arraystretch}{1.13}
\begin{tabularx}{\linewidth}{@{}>{\raggedright\arraybackslash}p{0.37\linewidth} X@{}}
\toprule
\textbf{Task / environment suffix} & \textbf{Objective and coordination requirement} \\
\midrule
\textbf{BendPick}\newline
{\texttt{BendPickMP-v0}}
& Bend toward a low table and grasp the target. The benchmark requires lifting it more than 5\,cm above the table, coupling body posture with reaching. \\
\addlinespace[4pt]
\textbf{Handover}\newline
{\texttt{HandoverTeleop-v0}}
& Transfer an object between the two hands and place it in a container, requiring coordinated grasping, handover, and release. \\
\addlinespace[4pt]
\textbf{Mobile P\&P}\newline
{\texttt{LocomotionPick\allowbreak BetweenTables\allowbreak Teleop-v0}}
& Pick up an object and transport it to a container on another table, maintaining the grasp during locomotion and completing placement. \\
\addlinespace[4pt]
\textbf{Grasp}\newline
{\texttt{TabletopGraspMP-v0}}
& Grasp a tabletop target and lift it more than 5\,cm above the table. This setting emphasizes local hand--object alignment and lifting. \\
\addlinespace[4pt]
\textbf{XMoveBendPick}\newline
{\texttt{XMoveBendPick\allowbreak Teleop-v0}}
& Move laterally, bend toward a low target, and lift it, combining body repositioning with a change in reaching height. \\
\addlinespace[4pt]
\textbf{XMovePick}\newline
{\texttt{XMovePickTeleop-v0}}
& Move laterally and lift a target object, testing the transition from locomotion to grasping. \\
\bottomrule
\end{tabularx}
\end{table}

\clearpage
\subsection{Task Demonstration Snapshots}
\label{app:simple_rollouts}
\begin{figure}[htbp]
    \centering
    \includegraphics[width=\linewidth]{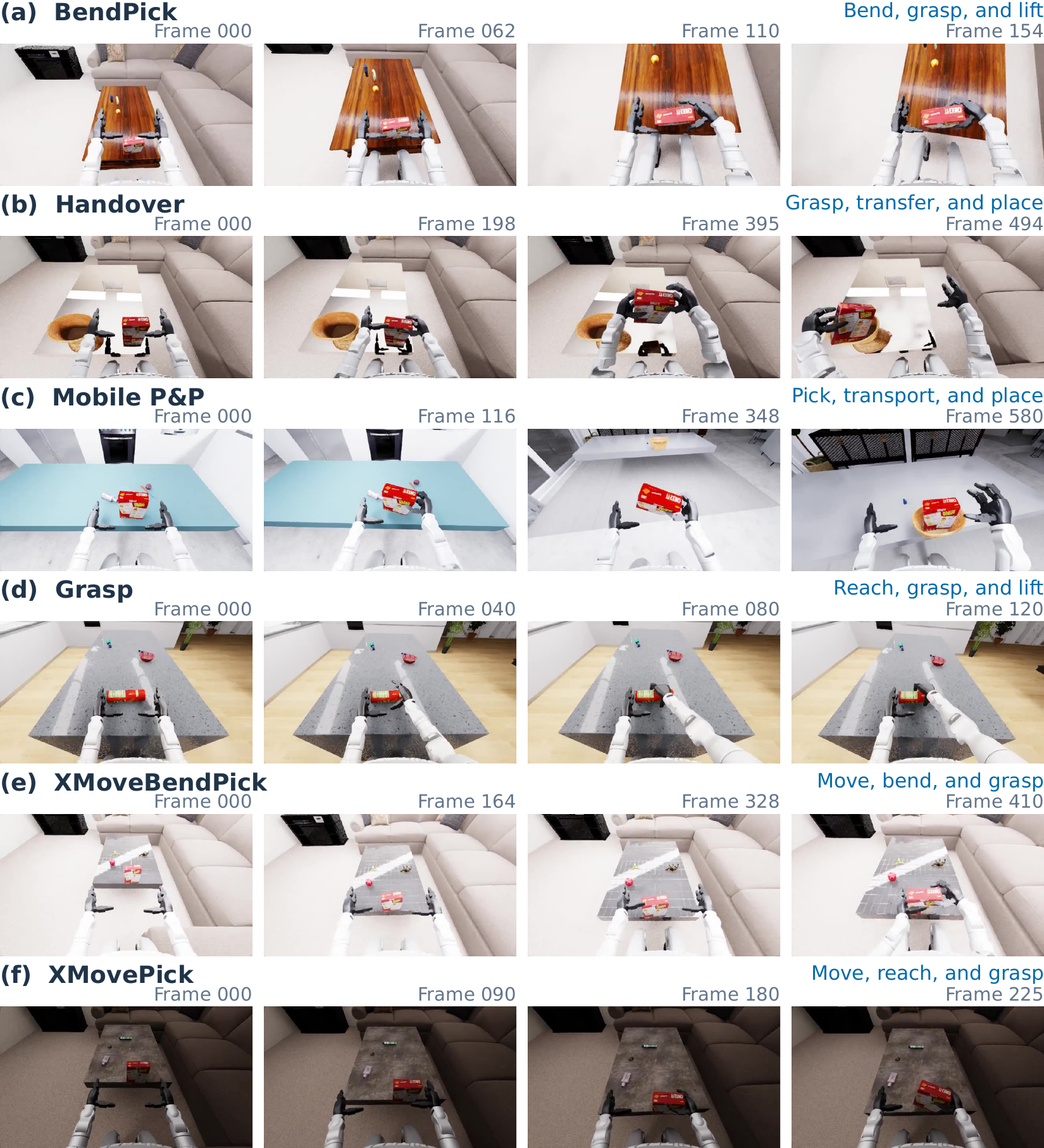}
    \caption{\textbf{Egocentric demonstrations of the six SIMPLE tasks.} Each row shows one dataset trajectory, with time advancing from left to right. BendPick and Grasp illustrate reaching and lifting; Handover shows bimanual transfer and placement; Mobile P\&P shows object transport between tables; and XMoveBendPick and XMovePick combine body repositioning with grasping. Frames retain their original field of view and colors. Frame indices refer to the source image sequences; the selected intervals may differ within and across tasks.}
    \label{fig:simple_task_rollouts}
\end{figure}

Figure~\ref{fig:simple_task_rollouts} illustrates the six tasks using one demonstration sequence per task from the SIMPLE dataset~\citep{wei2026simple}. Each row contains four selected egocentric observations showing the progression of the task, including reaching, object interaction, and, where applicable, transport or handover. Labels give the original frame indices within each sequence.

\paragraph{Discussion of Task Interference.}
Compared with language-only context, the full model improves substantially on XMovePick and BendPick but underperforms on XMoveBendPick. These tasks share similar visual observations (Figure~\ref{fig:simple_task_rollouts}), which may introduce task interference under joint multi-task training. Increasing model capacity may help alleviate this interference.

\clearpage
\subsection{Visual Variation Across Evaluation Levels}
\label{app:simple_visual_variation}
\begin{figure}[htbp]
    \centering
    \includegraphics[width=\linewidth,height=0.82\textheight,keepaspectratio]{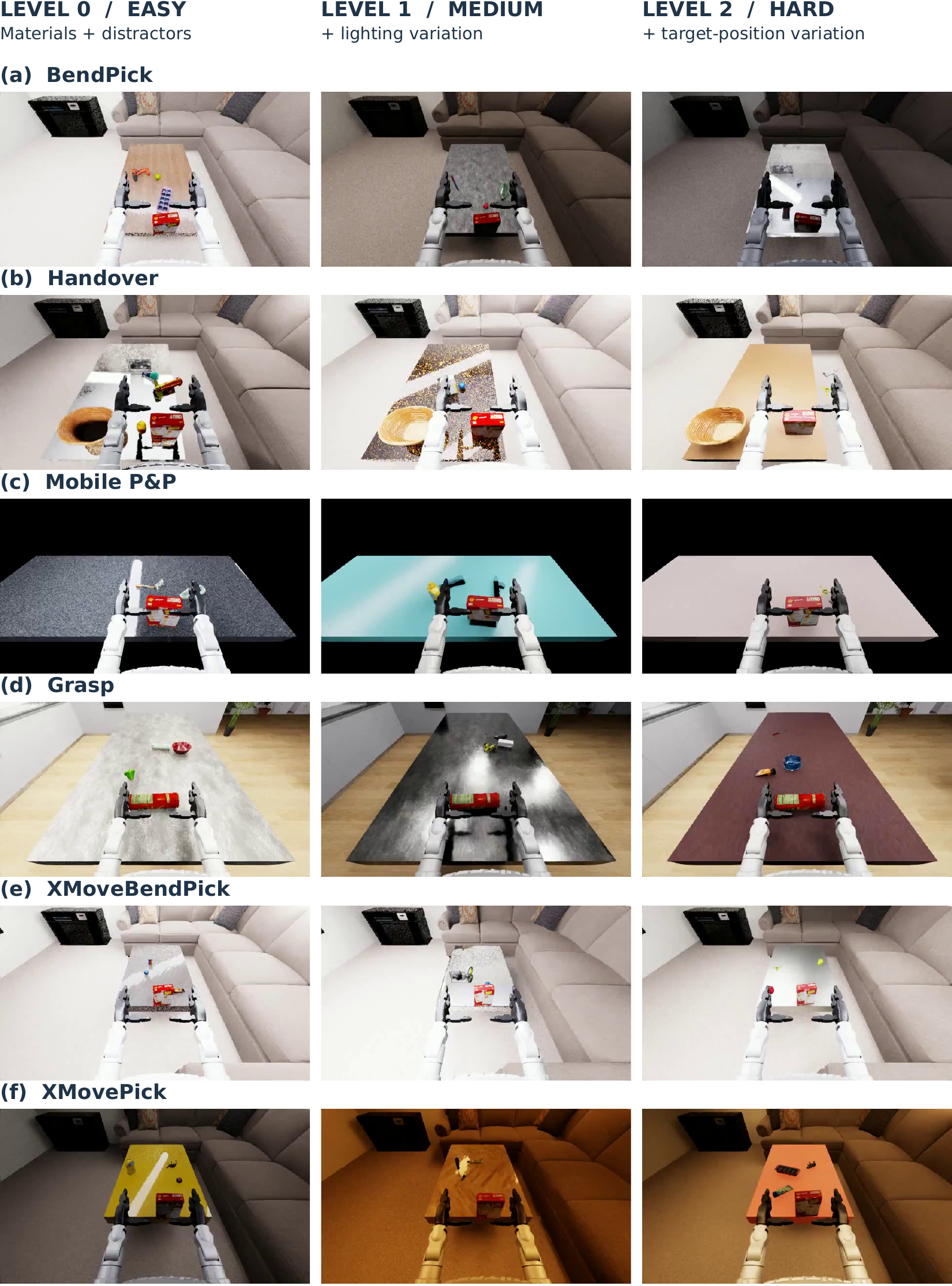}
    \caption{\textbf{Six tasks under three cumulative randomization levels.} Each cell is the first frame of episode~0 in the corresponding GR00T-N1.6 evaluation folder, viewed through the left head camera. Columns follow the level definitions in Table~\ref{tab:simple_levels}. These are separate evaluation scenes; the shared episode index does not establish matched random seeds across levels. Images retain the recorded lighting, materials, and object arrangements without color adjustment or cropping.}
    \label{fig:simple_randomization_levels}
\end{figure}
\clearpage

\section{Real-World Experimental Details}

\subsection{Task Descriptions}
\label{app:real_exp}

We quantitatively evaluate three real-world loco-manipulation tasks: Walk to Fish (W2F), Walk to Mirror (W2M), and Walk to Doll (W2D). All three tasks include approaching the target and performing the subsequent manipulation. We evaluate Being-M0.7, GR00T-N1.6, $\Psi_0$, and $\pi_{0.5}$ over five trials per task, totaling fifteen trials per method, as reported in Table~\ref{tab:real_world_all}. Physical markers fix the robot starting positions and the table positions across methods. Target objects are randomly initialized at locations spanning the left, center, and right regions of the task workspace. Each trial allows at most 30 seconds for the complete approach-and-manipulation sequence. A human evaluator judges success: the robot must successfully grasp the target doll in W2M and W2D, or scoop up the toy fish with the net in W2F. Trials that do not meet the corresponding criterion within 30 seconds are counted as failures.

\begin{figure}[ht]
    \centering
    \includegraphics[width=1.0\textwidth]{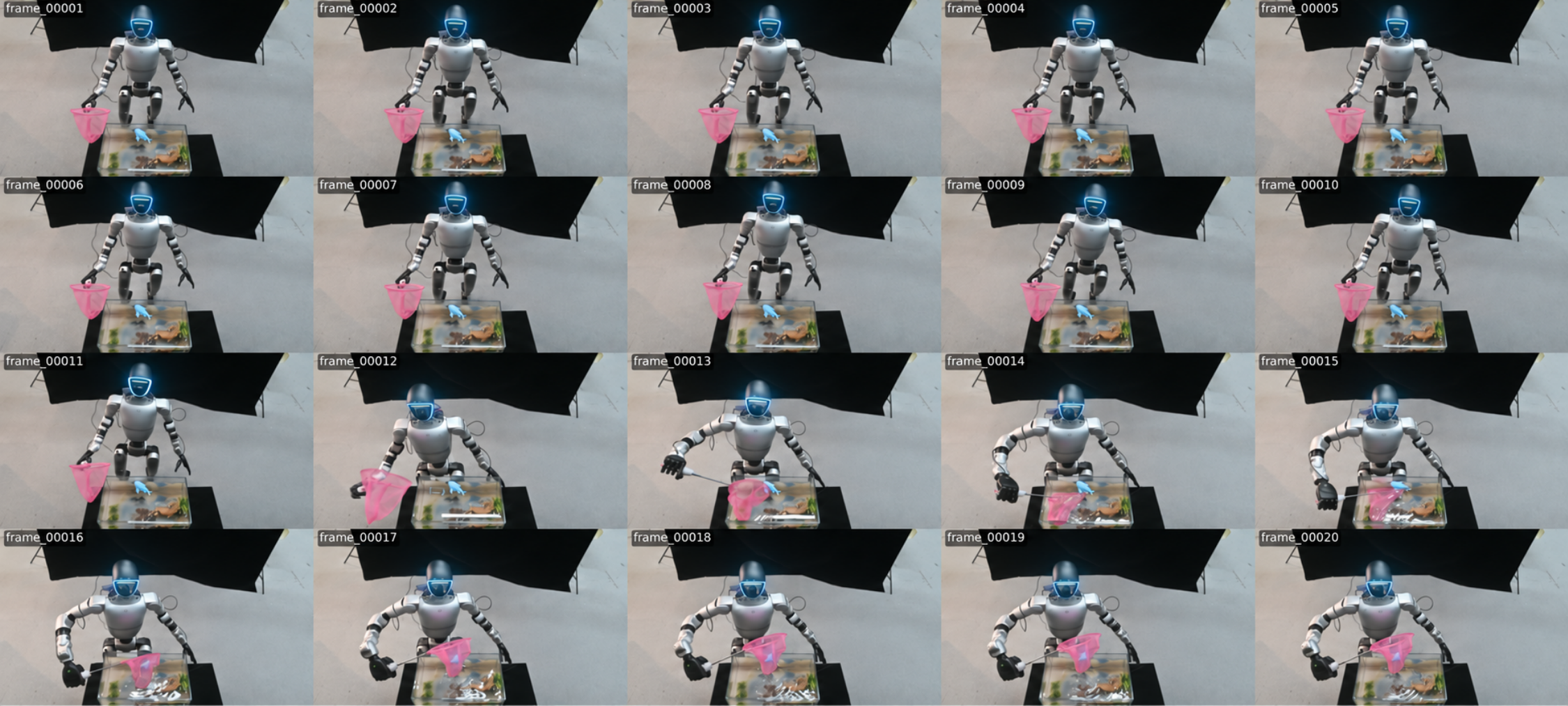} 
    \caption{Snapshots of the complete Walk to Fish task sequence.}
    \label{fig:fish_all}
\end{figure}

\paragraph{Walk to Fish (W2F).}
The robot walks toward a water tank, positions its body for interaction, and uses a handheld net to scoop up a toy fish. The quantitative evaluation covers this complete approach-and-scooping sequence. The task requires coordinated locomotion, body positioning, and tool use guided by visual feedback. The Water-Tank Fish Scooping case study in Figure~\ref{fig:fish_all} illustrates this sequence.

\begin{figure}[!htb]
    \centering
    \includegraphics[width=1.0\textwidth]{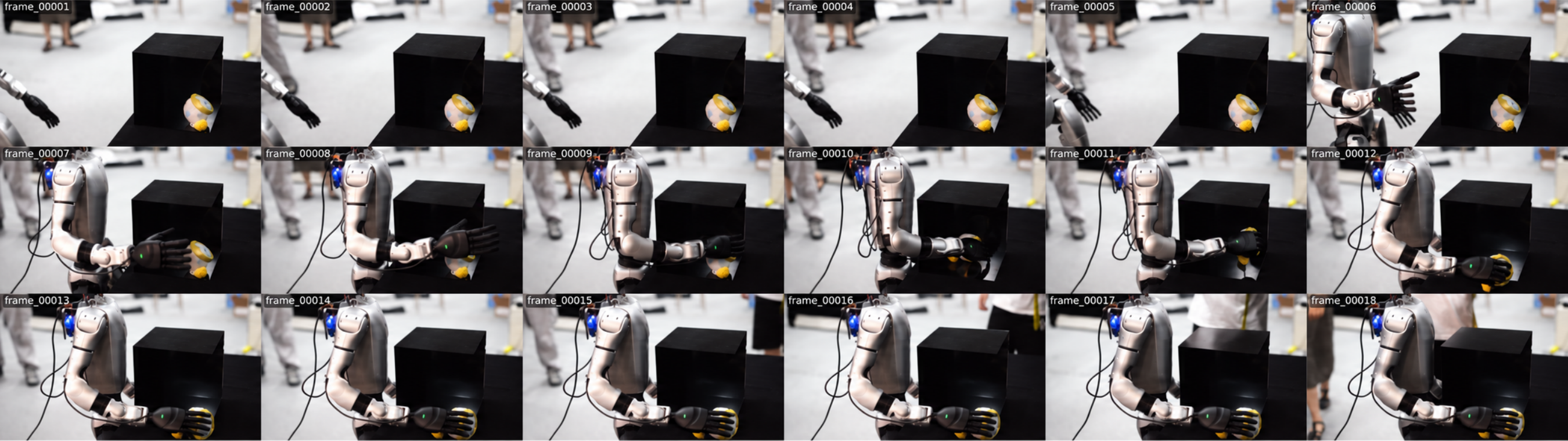} 
    \caption{Snapshots of the complete Walk to Mirror task sequence.}
    \label{fig:mirror_all}
\end{figure}

\clearpage
\paragraph{Walk to Mirror (W2M).}
The robot walks toward a table and approaches a partially enclosed box containing a toy. Since the toy is initially hidden from direct view, the robot must infer its location from a mirror reflection before reaching into the box to grasp it. The task combines indirect visual observations with whole-body positioning and manipulation. The Mirror Toy Grasping case study in Figure~\ref{fig:mirror_all} illustrates this behavior.

\paragraph{Walk to Doll (W2D).}
The robot walks toward a table and grasps a directly visible doll. This task evaluates the transition from locomotion to target-directed grasping, requiring the robot to coordinate its approach, body positioning, and arm motion. Together, W2M and W2D assess approach-and-grasp sequences with mirror-based and direct target observations, respectively.

\begin{figure}[!htb]
    \centering
    \includegraphics[width=1.0\textwidth]{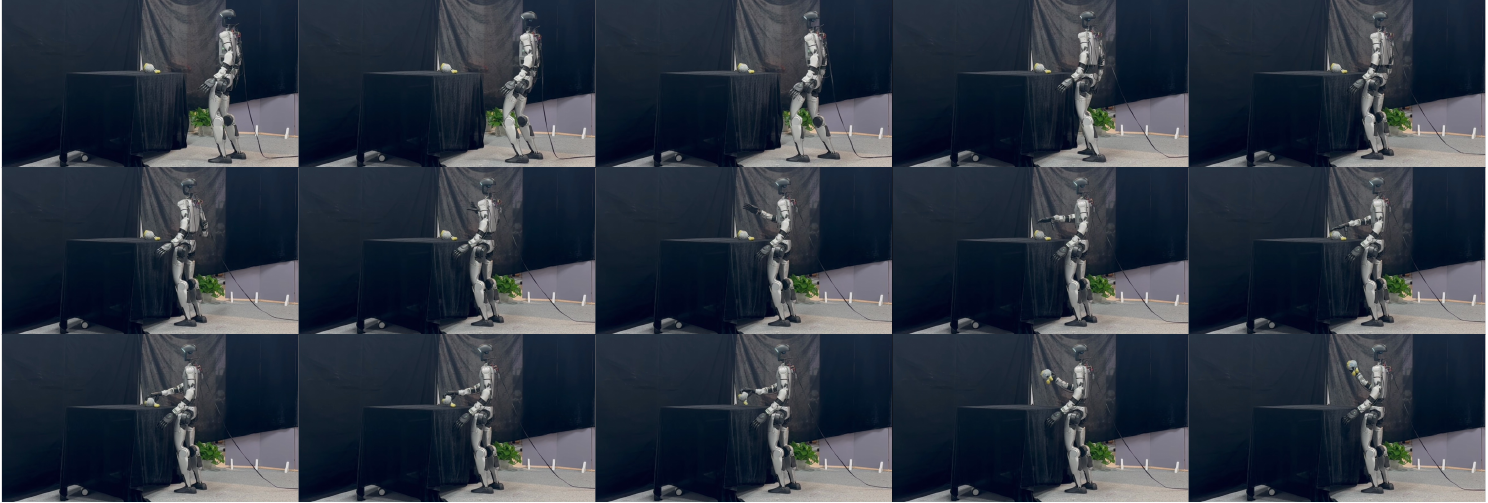} 
    \caption{Snapshots of the complete Walk to Doll task sequence.}
    \label{fig:doll_all}
\end{figure}

\end{document}